\documentclass[twocolumn]{svjour3}          
\smartqed  
\usepackage{natbib}
\usepackage{graphicx}

\usepackage[utf8]{inputenc} 
\usepackage[T1]{fontenc}    
\usepackage{url}            
\usepackage{booktabs,multirow,array}
\usepackage{amsfonts}       
\usepackage{nicefrac}       
\usepackage{microtype}      
\usepackage{graphicx}
\usepackage{lipsum}
\usepackage{natbib}
\usepackage{mathtools} 
\usepackage{amsmath}
\usepackage{svg}
\usepackage{extarrows}
\usepackage[]{threeparttable}

\usepackage{xcolor}
\usepackage{amssymb}
\usepackage[misc]{ifsym}

\usepackage[switch,pagewise]{lineno}

\usepackage[]{svg}
\usepackage{lmodern}

\begin{document}
\title{Wide-angle Image Rectification: A Survey \thanks{This work was supported by Australian Research Council Projects FL-170100117, IH-180100002, IC-190100031.} 
}

\authorrunning{Jinlong Fan, et al.} 
 
\author{Jinlong Fan$^{1}$,
        Jing Zhang$^{1}$,
        Stephen J. Maybank$^{2}$,
        Dacheng Tao$^{1}$
    }

\institute{
Jinlong Fan (jfan0939@uni.sydney.edu.au) \\
Jing Zhang (jing.zhang1@sydney.edu.au) \\
Stephen J. Maybank (sjmaybank@dcs.bbk.ac.uk) \\
\Letter~Dacheng Tao (dacheng.tao@sydney.edu.au)\\
$^{1}$ School of Computer Science,
Faculty of Engineering, The University of Sydney, Darlington,
NSW 2008, Australia.\\
$^{2}$ 
Department of Computer Science and Information System, Birkbeck College, University of London, U.K.
}

\date{Received: date / Accepted: date}

\maketitle

\begin{abstract}
Wide field-of-view (FOV) cameras, which capture a larger scene area than narrow FOV cameras, are used in many applications including 3D reconstruction, autonomous driving, and video surveillance. However, wide-angle images contain distortions that violate the assumptions underlying pinhole camera models, resulting in object distortion, difficulties in estimating scene distance, area, and direction, and preventing the use of off-the-shelf deep models trained on undistorted images for downstream computer vision tasks. Image rectification, which aims to correct these distortions, can solve these problems. In this paper, we comprehensively survey progress in wide-angle image rectification from transformation models to rectification methods. Specifically, we first present a detailed description and discussion of the camera models used in different approaches. Then, we summarize several distortion models including radial distortion and projection distortion. Next, we review both traditional geometry-based image rectification methods and deep learning-based methods, where the former formulates distortion parameter estimation as an optimization problem and the latter treats it as a regression problem by leveraging the power of deep neural networks. We evaluate the performance of state-of-the-art methods on public datasets and show that although both kinds of methods can achieve good results, these methods only work well for specific camera models and distortion types. We also provide a strong baseline model and carry out an empirical study of different distortion models on synthetic datasets and real-world wide-angle images. Finally, we discuss several potential research directions that are expected to further advance this area in the future.
\end{abstract}

\section{Introduction}  \label{intro}

\begin{figure*}[ht]
    \centering
    \includegraphics[width=0.95\linewidth]{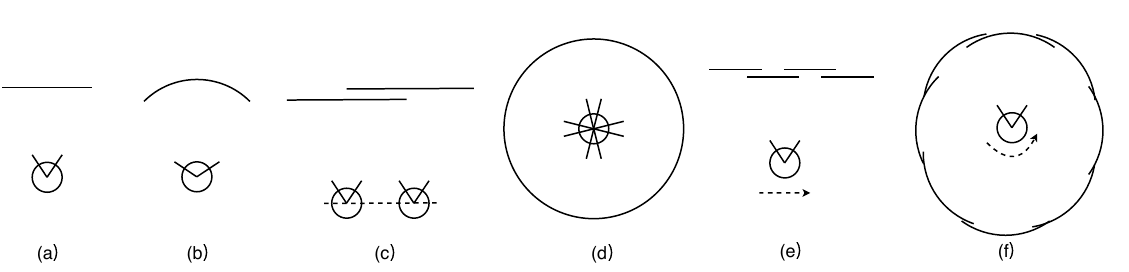}
    \caption{Illustration of different camera systems. (a) Conventional camera with a narrow FOV. (b) Wide FOV camera. (c) Stereo vision system. (d) Multi-view camera system. (e) Monocular wide FOV camera system via translation. (f) Monocular wide FOV camera system via rotation.}
    \label{fig:diffcameras}
\end{figure*}

Cameras efficiently capture dense intensity and color information in a scene and are widely used in different computer vision tasks including 3D reconstruction, object detection and tracking \citep{10.1007/s11263-007-0075-7,10.1007/s11263-009-0275-4,chen2020recursive}, semantic segmentation, and visual location and navigation \citep{10.1007/s11263-006-0023-y}. Not like some animal eyes that have a wide field-of-view (FOV) \citep{landAnimalEyes2012}, as the digital eyes of computers, normal cameras often have limited FOV, e.g., the most widely used monocular pinhole camera, which obeys the perspective transformation and linear projection rules, has a narrow FOV as illustrated in Figure~\ref{fig:diffcameras}(a). But the FOV of a camera system can be increased in different ways to capture more contents and facilitate visual analysis. For instance, a stereo vision system can be devised by leveraging two (identical) cameras spaced a certain distance apart to increase the FOV, as shown in Figure~\ref{fig:diffcameras}(c). Moreover, more than two cameras can be easily integrated into one visual system in some designed pattern for a larger or even $360^\circ$ FOV by overlapping the FOVs of neighboring cameras, as shown in Figure~\ref{fig:diffcameras}(d). Conversely, instead of using multiple cameras, a single camera with a narrow FOV can be moved (e.g., through yaw or pitch axis rotation or translation, as shown in Figure~\ref{fig:diffcameras}(e)-(f)) to cover a wide field from several frames.

Using multiple cameras or moving a single camera to obtain a large FOV requires extra processing (e.g., through camera calibration and point matching) to stitch spatially or temporally adjacent frames (i.e., panorama stitching),  which is computationally inefficient and challenging, especially in dynamic scenes or textureless areas. As an alternative, wide FOV cameras can achieve a large FOV using special lenses or structures, as shown in Figure~\ref{fig:diffcameras}(b). The most commonly used wide FOV camera is the omnidirectional camera, whose FOV covers a hemisphere or $360^\circ$ in the horizontal plane
\citep{nayar_CatadioptricOmnidirectionalCamera_1997, scaramuzza_OmnidirectionalCamera_2014}; fisheye and catadioptric cameras are two typical omnidirectional camera types. The fisheye camera (or dioptric camera) is a conventional camera combined with a shaped lens, while the catadioptric camera is equipped with a shaped mirror and lens \citep{yagi_PanoramaSceneAnalysis_1990,geyer_CatadioptricProjectiveGeometry_2001}. The fisheye camera has a FOV of approximately $180^\circ$ or more in the vertical plane, while the catadioptric camera has a $100^\circ$ FOV or more in the vertical plane. A catadioptric camera equipped with hyperbolic, parabolic, or elliptical mirrors is known as a central catadioptric camera \citep{baker_TheorySingleViewpointCatadioptric_1999}, which has only one effective viewpoint. The central camera has two attractive properties. First, the capturing distorted image can be geometrically corrected to a perspective image, since every pixel in the image corresponds to one particular incoming ray passing through the single viewpoint at a particular angle, which is measurable and can be derived after the camera calibration. Second, all central cameras follow the rigorous epipolar geometry constraint, which is well studied in multi-view vision. However, catadioptric cameras are complex and fragile due to their mirrors, so fisheye cameras are more popular in practice. For clarity, the term wide FOV camera in this paper includes central catadioptric cameras, fisheye cameras, and wide-angle cameras with radial distortion (normally a FOV < $120^\circ$). The image captured by the wide FOV camera is called a wide-angle image.

Wide FOV cameras can record more (or even all) visual contents in the scene via a single shot, which is very useful in many vision tasks, such as video surveillance, object tracking, simultaneous localization and mapping (SLAM) \citep{yagi_RealtimeOmnidirectionalImage_1994,rituerto_VisualSLAMOmnidirectional_2010,caruso_LargescaleDirectSLAM_2015,paya_StateoftheArtReviewMapping_2017,matsuki_OmnidirectionalDSODirect_2018}, structure from motion (SfM)  \citep{pengchang_OmnidirectionalStructureMotion_2000,scaramuzza_FlexibleTechniqueAccurate_2006,neumann_EyesEyesNew_2002}, and augmented reality/virtual reality (AR/VR) \citep{110003210133}. Wide FOV cameras can see more context and capture larger objects, making object tracking more stable \citep{posada_FloorSegmentationOmnidirectional_2010,markovic_MovingObjectDetection_2014} and detectors more effective \citep{cinaroglu_DirectApproachObject_2016,yang_ObjectDetectionEquirectangular_2018a}.

Although wide FOV cameras are useful, they break the perspective transformation relationship between real points and those in the image, resulting in distortions in the wide-angle image. These distortions make it hard to estimate distance, area, and direction and prevent taking wide-angle images directly as inputs to the off-the-shelf deep neural models trained on distortion-free images due to the explicit domain gap. To address this issue, wide-angle image rectification as an important vision task has been studied for decades and is still an active research area in the deep learning era. It aims to rectify the distortions in the wide-angle image to obtain an undistorted image obeying perspective transformation. Generally, distortion can be represented as extra intrinsic parameters in camera models (Section \ref{cameramodels}) or as separate and independent distortion parameters in distortion models (Section \ref{distortionmodels}).

Basically, there are two main groups of approaches for wide-angle image rectification. One group is the calibration-based methods, which try to estimate the intrinsic and extrinsic parameters of the camera model representing how a point in the 3D world is mapped to a corresponding point on the image plane. This process is also known as camera calibration, where distortion parameters can be estimated as a part of the intrinsic parameters of a wide FOV camera \citep{heikkila_FourstepCameraCalibration_1997}.

Camera calibration has a long history \citep{Brown71close-rangecamera, zhang_FlexibleNewTechnique_2000}, and detailed reviews and comparisons of different calibration methods can be found in \citep{caprile_UsingVanishingPoints_1990,clarke_DevelopmentCameraCalibration_1998}. The calibration of stereo vision systems \citep{sid-ahmed_DualCameraCalibration_1990,gennery1979stereo} and moving camera systems \citep{maybank_TheorySelfcalibrationMoving_1992} have also been studied, as the calibration of fisheye cameras \citep{shah_SimpleCalibrationProcedure_1994,shah_IntrinsicParameterCalibration_1996,swaminathanNonmetricCalibrationWideangle2000, kannala_GenericCameraCalibration_2004} and omnidirectional cameras \citep{swaminathanNonSingleViewpointCatadioptric2006, mei_SingleViewPoint_2007}.

Generic methods for more than one type of camera have also been proposed. For example, \cite{kannala_GenericCameraModel_2006} studied a generic camera model and calibration method for both conventional and wide FOV cameras, while \cite{urban_ImprovedWideangleFisheye_2015} reported a new calibration procedure for wide FOV cameras based on a comprehensive performance evaluation across several datasets. Once the camera parameters (including the distortion parameters) are calibrated, they can be used to rectify the wide-angle image according to the camera model, as surveyed in \citep{hughes_ReviewGeometricDistortion_2008,puig_CalibrationOmnidirectionalCameras_2012,zhang_SurveyCatadioptricOmnidirectional_2013}.

The other group of methods estimates the distortion parameters in the camera model or distortion model from the wide-angle image passively. Not like the active calibration methods, pre-designed chess boards or other pre-defined subjects are not necessary for estimation methods. In this survey, we mainly focus on this group of methods. Readers who care more about calibration-based methods could refer to the surveys mentioned above. In this paper, we comprehensively review the progress in this area from the fundamentals, including camera models and distortion models, to image rectification methods, including both traditional geometry-based methods and the more recent deep learning-based methods. Specifically, we first present a detailed description and discussion of the camera models, especially the wide FOV camera model. Then, we summarize several typical distortion models, including radial distortion models and projection distortion models. Next, we comprehensively review both traditional geometry-based image rectification methods and deep learning-based methods that estimate the distortion parameters (or equivalent warp field) in the camera model or the distortion model. We categorize the geometry-based methods into three groups: line-based methods, content-aware methods, and multi-view methods. The learning-based methods are categorized into two groups: model-based methods and model-free methods, based on whether a specific parameterized camera model or distortion model is leveraged in the framework. We further evaluate the performance of state-of-the-art methods and discuss their strengths and limitations. Moreover, we also establish a strong baseline and carry out an empirical study of different distortion models on both synthetic datasets and real-world wide-angle images. Finally, we discuss several research directions that might provide a more general solution.

Before going deeper, let us clarify some terminologies that may be confused first. As shown in Figure \ref{fig:pinhole}, the word with a subscript 'u' means the concept is in the undistorted image domain, while the word with a subscript 'd' means the distorted image domain. Warp field, flow field, and displacement field are generally replaceable in this survey. All of them mean the per-pixel field that represents the transformation between distorted images and undistorted ones. Camera models and distortion models are mathematical models that describe the distortion, while deep models mean the trained deep neural networks. But we may use the name of the distortion model that the training data is based on to call that deep model. For example, if a deep model is trained on a dataset that is synthesized under the $X$ distortion model, we may also name the deep model as $X$ model. But it is easy to decide the model is a deep model or a distortion model in context in Section \ref{performance}.

To our best knowledge, this is the first survey of wide-angle image rectification. The main contributions of this paper are as follows:
 \begin{itemize}
     \item we comprehensively describe and discuss the typical camera models and distortion models that are leveraged in most wide-angle image rectification approaches;

     \item we comprehensively review of the state-of-art methods for wide-angle image rectification, including traditional geometry-based and deep learning-based image rectification methods;
     
     \item we evaluate the performance of state-of-the-art methods and discuss their strengths and limitations on both synthetic datasets and real-world images and also propose a strong baseline model;
     
     \item we provide some insights into current research trends to highlight several promising research directions in the field.
 \end{itemize}

The rest of this paper is organized as follows. We first introduce several typical camera models and distortion models in Section~\ref{cameramodels} and  \ref{distortionmodels}. Details of the traditional geometry-based and learning-based methods are presented in Section~\ref{rectification}, followed by the performance evaluation in Section~\ref{performance}. Next, we provide some insights on recent trends and point out several promising research directions in this field in Section~\ref{future}. Finally, the concluding remarks are made in Section~\ref{conclusion}.

\section{Camera Models}  \label{cameramodels}  
Before the introduction of camera models, we first define the notations used in this paper. We use lowercase letters to denote scalars, e.g., $x$, bold lowercase letters to denote vectors, e.g. $\mathbf{f}$, and bold uppercase letters to denote matrices, e.g., $\mathbf{F}$. We use $\mathbf{w}=[X,Y,Z]^\mathsf{T} \in \mathbf{\Psi} \subset \mathbb{R}^3$ to represent a point in the 3D world coordinate $\mathbf{\Psi}$, $\mathbf{c}=[x,y,z]^\mathsf{T} \in \mathbf{\Omega} \subset \mathbb{R}^3$ to represent a point in the camera coordinate $\mathbf{\Omega}$, and $\mathbf{m}=[u, v]^\mathsf{T} \in \mathbf{\Phi} \subset \mathbb{R}^2$ to represent a pixel on the image plane $\mathbf{\Phi}$. Besides, we use a calligraphic uppercase letter to represent a mapping function, e.g., $\mathcal{M}$.

Camera model describes the imaging process between a point in the 3D world coordinate to its projection on the 2D image plane using a mathematical formulation. Different kinds of lens correspond to different kinds of camera models \citep{sturm_CameraModelsFundamental_2010}. Let $[X,Y,Z]^\mathsf{T}$ denote a point in the 3D world coordinate and $[u, v]^\mathsf{T}$ denote its corresponding point on the image plane. Camera model defines a mapping $\mathcal{M}$ between $[X,Y,Z]^\mathsf{T}$ and $[u, v]^\mathsf{T}$:
\begin{equation}
[u,v]^\mathsf{T} = \mathcal{M} (X,Y,Z),
\label{eq:projection}
\end{equation}

or in the homogeneous form:
\begin{equation}
[u,v,1]^\mathsf{T} = \mathcal{M} (X,Y,Z,1). 
\label{eq:projection_homo}
\end{equation}
Generally, the projection can be divided into four steps:
\begin{enumerate}
    \item In the first step, the 3D point $[X, Y, Z]^\mathsf{T}$ is transformed to the camera coordinate via a $3\times3$ rotation $\mathbf{R}$ and a $3$-dimension translation $\mathbf{t}$, i.e.,
    \begin{equation}
        [x_c, y_c, z_c]^\mathsf{T} = [\mathbf{R}|\mathbf{t}][X, Y, Z, 1]^\mathsf{T},
    \end{equation}
    where $[x_c, y_c, z_c]^\mathsf{T}$ is the corresponding point in the camera coordinate. The $3 \times 4$ matrix $[\mathbf{R}|\mathbf{t}]$ is called extrinsic camera matrix. This step is a rigid transformation and no distortion is involved.
    
    \item In the second step, the point $[x_c, y_c, z_c]^\mathsf{T}$ is projected onto a surface, which could be a plane or not. In the pinhole camera model, this surface is a plane at $z = 1$, and the normalized coordinate is $[x_n, y_n]^\mathsf{T} = [\frac{x_c}{z_c}, \frac{y_c}{z_c}]^\mathsf{T}$. But in most wide FOV camera models, this surface is normally a quadratic one. The points on this projection surface are then normalized to $z=1$. Here we can use a transformation function $\mathcal{N}$ to denote this normalization:
    \begin{equation}
    \left\{\begin{matrix}
    x_n &= \frac{x_c}{\mathcal{N}(x_c, y_c, z_c)}
    \\ 
    y_n &= \frac{y_c}{\mathcal{N}(x_c, y_c, z_c)}
    \end{matrix}\right.,   
    \label{eq:normalization}
    \end{equation}
    \item In the third step, other types of distortions may be introduced to represent the displacement caused by the manufacturing defect or wide-angle lens. The distortions can be mathematically described by a specific distortion model. In a practical application, one specific type of distortion model is usually used in one camera model. More details will be presented in Section~\ref{distortionmodels}. Given the mapping function of the distortion model $\mathcal{D}$, the distorted image coordinate $[x_d, y_d]^\mathsf{T}$ is formulated as:
    \begin{equation}
    \left\{\begin{matrix}
    x_d &= \mathcal{D}(x_n)
    \\ 
    y_d &= \mathcal{D}(y_n)
    \end{matrix}\right.,   
    \label{eq:distorted}
    \end{equation}
    \item In the final step, the distorted point on the normalized plane is projected onto the image plane via a $3 \times 3 $ intrinsic camera matrix $\mathbf{K}$:
    \begin{equation}
        [u, v, 1]^\mathsf{T} = \mathbf{K} [x_d, y_d, 1]^\mathsf{T},
    \label{eq:projection_K}
    \end{equation}
    \begin{equation}
    \mathbf{K} \triangleq \begin{bmatrix} f_x m_u & s & u_0 \\ 
                                          0 & f_y m_v & v_0 \\
                                          0 & 0 & 1\end{bmatrix},
    \end{equation}
    where $f_x$ and $f_y$ are the focal length at x and y axis, respectively. In most cases, they are the same and denoted as $f$. $s$ is the skew parameter. If $x$-axis and $y$-axis are perpendicular to each other, $s$ is zero. $m_u$ and $m_v$ are the number of pixels per unit distance in $u$ and $v$ direction, respectively. If $m_u$ is the same as $m_v$, the camera has square pixels. $[u_0, v_0]^\mathsf{T}$ is the coordinate of the image center. For most cameras, we can set $s=0, m_u=m_v$ and focal length in pixel unit, then Eq.~\eqref{eq:projection_K} can be re-written as:    
    \begin{equation}
    \left\{\begin{matrix}
    u &= f_x x_d + u_0
    \\ 
    v &= f_y y_d + v_0
    \end{matrix}\right.,   
    \label{eq:mapuv}
    \end{equation}
    which is a linear transformation that keeps shapes.
\end{enumerate} 

The first step and the last step are almost the same for different camera models, which are distortion-free. By contrast, the second step and the third step are crucial for accurately representing wide FOV cameras and distortions. So, when we introduce camera models, we will focus on the imaging process in these two steps. We are not going to collect all the camera models in this survey. Instead, only the ones that are most commonly used in the computer vision community, especially for wide FOV cameras, will be introduced, i.e., the pinhole camera model (PCM), the unified camera model (UCM), the extended unified camera model (EUCM), and the double sphere camera model (DSCM), as summarized in Table~\ref{tab:cameradistortion}. Details of these models are presented as follows.

\begin{table}[ht]
\scriptsize
  \centering
    \caption{Typical camera models for wide FOV cameras.}  
    \label{tab:cameradistortion}
    \begin{tabular}{ c l}
    \toprule
    Camera model & Wide FOV Camera\\
    \midrule
    PCM & perspective camera, wide-angle camera \\
    \midrule
    {UCM} & central catadioptric camera, fisheye camera \\
    \midrule
    {EUCM} & central catadioptric camera, fisheye camera \\
    \midrule
    {DSCM} & fisheye camera \\
    \bottomrule
    \end{tabular}
\end{table}

\subsection{Pinhole Camera Model} \label{pcm}
The pinhole camera model (PCM) is the most common and widely used camera model in computer vision. It can be seen as a first-order approximation of the conventional camera without geometric distortions. For conventional cameras, which have a small field of view (normally a FOV $< 90^{\circ}$) and obey the perspective transformation, this approximation is accurate enough. But for wide FOV cameras, the performance of PCM will degrade significantly.

\begin{figure}[ht]
    \centering
    \includegraphics[width=0.8\linewidth]{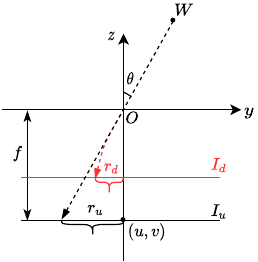}
    \caption{The pinhole camera model and distortion.} 
    \label{fig:pinhole} 
\end{figure}

The pinhole aperture in a pinhole camera is assumed to be an infinitely small point and all projection lines must pass through this point, i.e., it is a central camera and has one single effective viewpoint. As shown in Figure \ref{fig:pinhole}, $O$ is called the optical center and the line passes through the optical center perpendicular to the image plane $I_u$ is the optical axis, i.e., the $z$ axis of the camera coordinate. All points on the optical axis will project to the principal point on the image plane. In most cases, this principal point is the center of the image $[u_0, v_0]^\mathsf{T}$. The distance from the optical center to the principal point is the focal length $f$. A 3D point $W = [x_c, y_c, z_c]^\mathsf{T}$ in the camera coordinates projects onto the image plane as:
\begin{equation}
\left\{\begin{matrix}
u &= f_x \frac{x_c}{z_c} + u_0
\\ 
v &= f_y \frac{y}{z_c} + v_0
\end{matrix}\right..
\label{eq:pcm-1}
\end{equation}

Assuming there is an incident ray passing through $[x_c, y_c, z_c]^\mathsf{T}$ and optical center $O$ with an incident angle $\theta$ to the optical axis, the radial distance $r$ from the image point to the principal point can be calculated as:
\begin{equation}
    r = f \tan \theta. 
    \label{eq:ptheta}
\end{equation}
Here it is easy to find that $\theta$ should be smaller than $90^\circ$ (since FOV equals two times of $\theta$, so the FOV is smaller than $180^\circ$). Otherwise, the incoming ray will not intersect with the image plane, i.e., there is no projection point on the image plane, which means the pinhole camera can not see anything behind. Most cameras can not see all the points in the 3D world at one time because of the limited FOV. We define the points that could be projected onto the image plane in the camera model as the valid projection set and the projection of the point in the valid projection set is a valid projection. Thereby, The valid projection of PCM is defined on $\mathbf{\Psi} = \{\mathbf{w} \in \mathbb{R}^3 |z>0\}$. For a wide FOV camera with a FOV smaller than $120^\circ$, PCM can be used to describe the moderate distortions together with a proper distortion model (see Section \ref{distortionmodels}). However, when FOV becomes larger, e.g., FOV $> 120^\circ$, a wide FOV camera model could be a better choice for higher accuracy.

\subsection{Unified Camera Model} \label{unifiedmodel}
\begin{figure*}[ht]
    \center
    \includegraphics[width=0.75\textwidth]{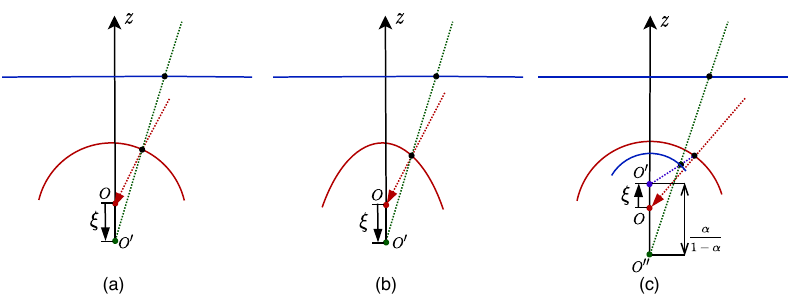}
    \caption{Illustration of the (a) unified camera model, (b) extended unifided caemra model, and (c) double sphere camera model. Red curves denote the projection surface. Blue lines denote the normalized image planes. The red lines with arrows denote the incident rays. $\xi$ is the distance from the original optical center $O$ to the new virtual one $O^{\prime}$.}
    \label{fig:3camera}
\end{figure*}

In PCM, the normalization function is $\mathcal{N}(x,y,z) = z$. When $z \rightarrow 0$, the accuracy of the model will drop dramatically, especially for large $r$. By contrast, the unified camera model(UCM) can work correctly when $z$ is zero or even negative, which means the FOV of the camera can be bigger than $180^\circ$. The normalization function in UCM is defined as \citep{geyer_UnifyingTheoryCentral_2000}:
\begin{equation}
    \mathcal{N}(x,y,z) = z + \xi r_s
    \label{eq:UCMnorm}
\end{equation}
where $r_s = \sqrt{x^2 + y^2 + z^2}$ and $\xi$ is a projection parameter. As shown in Figure \ref{fig:3camera}(a), in UCM, a point is first projected onto a unit sphere in red (the projection surface) and then onto the the normalization plane in blue. Note that in the second step, a virtual optical center is used by shifting from the original one of the PCM by a distance $\xi$. According to Eq. \ref{eq:normalization}, the point on the normalization plane can be calculated as:
\begin{equation}
    \mathbf{n} = [x_n, y_n, 1]^\mathsf{T} = [\frac{x_c}{z_c+\xi r_s}, \frac{y_c}{z_c+\xi r_s}, 1]^\mathsf{T}.
    \label{eq:normpts}
\end{equation}
UCM is the same as PCM if $\xi=0$. And the larger the $\xi$ is, the wider FOV the UCM can handle. For a conventional camera, $\xi$ is expected to be small, while for a wide FOV camera, e.g. fisheye camera, $\xi$ should be large. The valid projection of UCM is defined on $\mathbf{\Psi} = \{\mathbf{w} \in \mathbb{R}^3 |z>-\xi r_s\}$.

A slightly modified version of UCM was proposed in \citep{mei_SingleViewPoint_2007}, which can describe both radial distortion and tangential distortion, thereby better suited for real-world cameras. Besides, although UCM was initially proposed for central catadioptric cameras \citep{geyer_UnifyingTheoryCentral_2000}, it had been extended to fisheye camera later in \citep{ying_CanWeConsider_2004, barreto_UnifyingGeometricRepresentation_2006}. Moreover, the discussion about the equivalence of UCM to pinhole-based model and capturing rays-based model can be found in \citep{courbon_GenericFisheyeCamera_2007}.

\subsection{Extended Unified Camera Model}
In \citep{khomutenko_EnhancedUnifiedCamera_2016}, it was pointed out that the distortion in UCM is actually equivalent with the even order polynomial distortion model (see details in Section \ref{distortionmodels}). Motivated by this, an enhanced unified camera model (EUCM) was proposed, where the normalization function $\mathcal{N}$ is defined as:
\begin{align}
    \mathcal{N}(x,y,z) &= \alpha \rho + (1-\alpha)z,\\
    \rho &= \sqrt{\beta(x^2+y^2) + z^2}.
\end{align}
Here, $\alpha$ and $\beta$ are two projection parameters, subjected to $\alpha \in [0,1]$, $\beta > 0$, and $\alpha \rho + (1-\alpha)z > 0$. $\alpha$ defines the type of the projection surface and $\beta$ can be used to adjust the shape of the projection surface. When $\beta = 1$, EUCM degrades to UCM with $\xi=\frac{\alpha}{1-\alpha}$. The normalised point is calculated as: 
\begin{equation}
    \mathbf{n} = [x_n, y_n, 1]^\mathsf{T} = [\frac{x_c}{\alpha \rho + (1-\alpha)z}, \frac{y_c}{\alpha \rho + (1-\alpha)z}, 1]^\mathsf{T}.
    \label{eq:eucmnormpts}
\end{equation}
The valid projection of EUCM is defined as follows:
\begin{align}
    \mathbf{\Psi} &= \{\mathbf{w} \in \mathbb{R}^3 |z>-w \rho\},\\
    w &= \begin{cases} 
        	\frac{\alpha}{1-\alpha}, &if \alpha \leq 0.5 \\
        	\frac{1-\alpha}{\alpha}, &if \alpha > 0.5 
		  \end{cases}
\end{align}
As shown in Figure \ref{fig:3camera}(b), the projection surface of EUCM is an ellipsoid, rather than the sphere of UCM, which can describe large distortions in a wide FOV lens better.

\subsection{Double Sphere Camera Model}
In \cite{usenko_DoubleSphereCamera_2018}, a novel camera model named double sphere camera model (DSCM) was proposed, which is well-suited for fisheye cameras and makes a good trade-off between accuracy and computational efficiency. The normalization function in DSCM is defined as:
\begin{align}
    \mathcal{N}(x,y,z) &= \alpha d_2 + (1-\alpha)(\xi d_1 + z),\\
    d_1 &= \sqrt{x^2 + y^2 + z^2}, \label{eq:dscm_norm1}\\
    d_2 &= \sqrt{x^2 + y^2 + (\xi d_1 + z)^2}, \label{eq:dscm_norm2}
\end{align}
where $\xi$ and $\alpha$ are two projection parameters. In DSCM, a point is first projected onto two spheres sequentially, the centers of which are shifted by $\xi$, as shown in Figure \ref{fig:3camera}(c). Then, the point is projected onto the normalization plane shifted by $\frac{\alpha}{1-\alpha}$. Accordingly, the normalized point is calculated as: 
\begin{small}
\begin{align}\nonumber
    \mathbf{n} &= [x_n, y_n, 1]^\mathsf{T} \\
               &= [\frac{x_c}{\alpha d_2 + (1-\alpha)(\xi d_1 + z)}, \frac{y_c}{\alpha d_2 + (1-\alpha)(\xi d_1 + z)}, 1]^\mathsf{T}.
    \label{eq:dscm_norm_pts}
\end{align}
\end{small}
The valid projection of DSCM is defined as follows:
\begin{align}
    \mathbf{\Psi} &= \{\mathbf{w} \in \mathbb{R}^3 |z>-w_2 d_1\}, \\
    w_2 &= \frac{w_1 + \xi}{\sqrt{2w_1\xi + \xi^2 + 1}}, \\
    w_1 &= \begin{cases}
        	\frac{\alpha}{1-\alpha}, &if \alpha \leq 0.5 \\
        	\frac{1-\alpha}{\alpha}, &if \alpha > 0.5 
		  \end{cases}.
\end{align}

\section{Distortion Models} \label{distortionmodels}
The real-world lens always has some kinds of distortions due to the imprecise manufacture or the nature of the wide-angle lens. When we talk about distortions in an image, a standard undistorted image is usually assumed, which is taken by an ideal lens, i.e., the PCM. Then, analyzing and recovering images from the distortions can be done according to specific distortion models. The difference between the camera model and the distortion model is that the camera model describes how a point in the scene is projected onto the image plane, while the distortion model focuses on the relationship between the distorted point coordinate and the undistorted point coordinate, i.e., the mapping from a distorted image to an undistorted image. The camera model and distortion model could work independently or together when it is necessary.

The parameterized distortion model \citep{sturm_CameraModelsFundamental_2010} describes the mapping from a point $[x_d, y_d]^\mathsf{T}$ in the distorted image to that in the undistorted image $[x_u, y_u]^\mathsf{T}$, which is the target after rectification, i.e.,
\begin{equation}
    [x_u, y_u]^\mathsf{T} = \mathcal{F} (x_d, y_d; c_x, c_y, \Theta).
    \label{eq:distortion_model}
\end{equation}
Here, $\mathcal{F}(\cdot)$ represents the distortion model, $[c_x, c_y]^\mathsf{T}$ denotes the distortion center, and $\Theta$ is a set of distortion parameters. Rectification refers to estimating the parameters $\Theta$ of the distortion model. $[c_x, c_y]^\mathsf{T}$ can be set to the center of the image, which is a reasonable approximation in most cases \citep{weng_CameraCalibrationDistortion_1992}.

Two principal types of distortions are radial distortion and decentering distortion (one type of the tangential distortion) \citep{hugemann_CorrectingLensDistortions_2010}. Accordingly, the distortion model $\mathcal{F}(\cdot)$ can be parameterized as follows \citep{Brown71close-rangecamera,prescott_LineBasedCorrectionRadial_1997}:
\begin{equation}
 \left\{\begin{matrix}
    x_u &= x_d + \Bar{x}(k_1 r_d^2 + k_2 r_d^4 + k_3 r_d^6 + \cdots) \\ 
    &+ (p_1 (r_d^2 + 2\Bar{x}^2) + 2p_2\Bar{x}\Bar{y})(1+p_3r_d^2 + \cdots)
    \\ 
    y_u &= y_d + \Bar{y}(k_1 r_d^2 + k_2 r_d^4 + k_3 r_d^6 + \cdots) \\ 
    &+ (p_2 (r_d^2 + 2\Bar{y}^2) + 2p_1\Bar{x}\Bar{y})(1+p_3r_d^2 + \cdots)
    \end{matrix}\right.,   
    \label{eq:distortion_parameterization}
\end{equation}
\begin{align}
   \Bar{x} &= x_d - c_x, \\
   \Bar{y} &= y_d - c_y, \\
   r_d &= \sqrt{(x_d - c_x)^2 + (y_d - c_y)^2}.
   \label{eq:distortion_parameterization_xyr}
\end{align}
Here, $r_d$ is the radial distance from an image point to the distortion center. $(k_1, k_2, k_3, \dots)$ are the coefficients of the radial distortion, while $(p_1, p_2, p_3, \dots)$ are the coefficients of the decentering distortion. Note that the high-order terms of the distortion are insignificant compared to the low-order terms \citep{weng_CameraCalibrationDistortion_1992} and the tangential distortions in practical lens are small and negligible \citep{cucchiara_HoughTransformbasedMethod_2003, sturm_CameraModelsFundamental_2010}. Therefore, only radial distortions are considered in most literature.

\subsection{Radial Distortions} \label{section:radialdistortion}
Radial distortions are the main distortions in central single-view camera systems, which cause points on the image plane to be displaced from the ideal position projected under the perspective camera model along the radial axis from the center of the distortion \citep{hughes_ReviewGeometricDistortion_2008}. A typical feature of this type of distortion is circular symmetry to the distortion center. Distortion models that represent radial distortions can be seen as nonlinear functions of the radial distance. Many models are proposed in the literature to describe radial distortions, which can be divided into two groups \citep{courbon_GenericFisheyeCamera_2007,ying_ImposingDifferentialConstraints_2015}, i.e., pinhole-based models and capturing rays-based models.

\textbf{Pinhole-based models}: The first group of models is based on the pinhole camera model (PCM in Section \ref{pcm}). The coordinate of a distorted point on the image plane is directly transformed from the coordinate of the point projected via the perspective model. The radial distance $r$ from a point to the distortion center is used to link the transformation $T_1$:
\begin{equation}
    r_u \xleftrightarrow{T_1} r_d,
    \label{eq:pinhole_distortion_model}
\end{equation}
where $r_u=\sqrt{(x_u - c_x)^2+(y_u - c_y)^2}$ is the radial distance on the undistorted image plane and $r_d$ is the distance on the distorted image plane. Typically, two types of distortion models, i.e., the polynomial model and the division model, are mostly used in practice \citep{santana-cedres_InvertibilityEstimationTwoParameter_2015}. In \citep{tsai_VersatileCameraCalibration_1987,mallon_PreciseRadialUndistortion_2004,ahmed_NonmetricCalibrationCamera_2005}, an odd polynomial model was proposed, i.e.,
\begin{align}
    r_u &= r_d + \sum_{n=1}^\infty k_n r_d^{2n+1} \notag \\
        &= r_d + k_1 r_d^3 + k_2 r_d^5 + \cdots \\ 
        &= r_d (1 + k_1 r_d^2 + k_2 r_d^4 + \cdots) \notag.
    \label{eq:polynomial_model}
\end{align}
This distortion model can describe small distortions but are insufficient to describe large ones introduced by fisheye lens \citep{hughes_ReviewGeometricDistortion_2008}. Therefore, a more general polynomial model was proposed in \citep{shah_SimpleCalibrationProcedure_1994} by using both odd terms and even terms. Polynomial Fish-Eye Transform \citep{basu_AlternativeModelsFisheye_1995} also included a 0th order term for better capacity . The polynomial model can work well when the distortions are small but when the distortions become large, the number of parameters and the order of the model would increase rapidly, leading to extensive computational load, which makes it unsuitable in real applications. By contrast, the division model  \citep{fitzgibbon_SimultaneousLinearEstimation_2001} can handle large distortions using fewer parameters, which is often used in image rectification:
\begin{equation}
    r_u = \frac{r_d}{1 + k_1 r_d^2 + k_2 r_d^4 + \cdots}.
    \label{eq:division_model}
\end{equation}
In addition to these two commonly used distortion models, many other forms of distortion models are also introduced in the literature. The most important difference between these models is the form of the function that is used to describe the relationship between $r_u$ and $r_d$. For example, Fish-Eye Transform in \citep{basu_AlternativeModelsFisheye_1995} used logarithmic function, field-of-view model in \citep{devernay_StraightLinesHave_2001} linked $r_u$ and $r_d$ using trigonometric function, and rational function was used in \citep{Li05anon-iterative}. Some of the typical pinhole-based distortion models are summarized in Table \ref{tab:fisheye-R}.

\begin{footnotesize}
\begin{table}[ht]
    \centering
    \caption{Typical pinhole-based distortion models.More details can be found in \citep{sturm_CameraModelsFundamental_2010}.}
    \label{tab:fisheye-R}
    \begin{tabular}{ r l}
    \toprule
    \textbf{Model Name} & \textbf{Equation} \\
    \midrule
    Polynomial Radial & $r_u = r_d(1 + k_1 r_d^2 + ...)$ \\
    \hline
    Fish-Eye Transform \\ \citep{basu_AlternativeModelsFisheye_1995} & $r_d = s\ln(1 + \lambda r_u)$ \\
    \hline
    Poly. Fish-Eye Transform \\ \citep{basu_AlternativeModelsFisheye_1995} & $r_d = r_u \sum_{n=0}^{\infty} k_n r_u^{n}$ \\
    \hline
    Field-of-View \\ \citep{devernay_StraightLinesHave_2001} & $r_d = \frac{1}{\omega} \arctan(2 r_u \tan(\frac{\omega}{2}))$ \\
    \hline
    Typical Division & $r_u = \frac{r_d}{1 + k r_d^2}$ \\
    \hline
    Rational Model \\ \citep{Li05anon-iterative} & $r_d = r_u \frac{\sum_{i=1}^{N_1} k_i r_u^{2i}}{\sum_{j=1}^{N_2} k_j r_u^{2j}}$ \\
    \bottomrule
    \end{tabular}
\end{table}
\end{footnotesize}

\begin{table}[ht]
    \centering
    \caption{Typical capturing rays-based distortion models.}
    \label{tab:fisheye-theta}
    \begin{tabular}{ r l}
    \toprule
    Model Name & Equation \\
    \midrule
    Rectilinear/Perspective & $r_d = f \tan \theta$ \\
    Equi-solid angle & $r_d = 2f \sin(\theta\ / 2)$ \\
    Equidistant/-angular & $r_d = f \theta$ \\
    Stereographic & $r_d = 2 f \tan(\theta / 2)$ \\
    Orthographic/Sine law & $r_d = f \sin \theta$ \\
    Polynomial & $r_d = f(k_1 \theta + k_2 \theta^3 + k_3 \theta^5 + \dots)$ \\
    \bottomrule
    \end{tabular}
\end{table}

\textbf{Capturing rays-based models}: 
This kind of distortion model is based on the capturing rays, where the relationship $T_2$ between the radial distance on the distorted image $r_d$ and the incident angle $\theta$ is used:
\begin{equation}
    r_d \xleftrightarrow{T_2} \theta.
    \label{eq:capturedray_distortion_model}
\end{equation}

For a pinhole camera, the incident angle $\theta$ is mapped to distorted radial distance $r_d$ according to Eq.~\eqref{eq:ptheta}, which is called the rectilinear model or perspective model and is not valid anymore for wide FOV cameras. To address this issue, different capturing rays-based distortion models are proposed for wide FOV cameras, e.g., 1) the equidistant (a.k.a equiangular) model proposed in \citep{kingslake_HistoryPhotographicLens_1989}, which is suitable for cameras with limited distortions; 2) the stereographic model proposed in \citep{stevenson_NonparametricCorrectionDistortion_1996}, which preserves circularity and projects 3D local symmetries onto 2D local symmetries; 3) the orthogonal (a.k.a sine law) model in \citep{ray_AppliedPhotographicOptics_2002} 4) the equi-solid angle model proposed in \citep{miyamoto_FishEyeLens_1964}; 5) the polynomial model proposed in \citep{kannala_GenericCameraCalibration_2004}. These models and their mapping functions are summarized in Table \ref{tab:fisheye-theta}. It can be found that the equidistant model is a specific case of the polynomial model with $k_1=1$ and $k_{2,...,n}=0$. The perspective model and stereographic model both use the tangent function, while the equi-solid angle \citep{miyamoto_FishEyeLens_1964} model and the orthographic model use sine function. Furthermore, both the tangent function and sine function can be represented by a series of odd-order terms of $\theta$ using Taylor expansion, which has the same form as the polynomial model. Therefore, the polynomial model can be seen as a generalization of other models. The polynomial model can achieve high accuracy with adequate parameters, but the computation would be expensive. In real-life applications, it is often used with a fixed number of parameters, e.g., typically with five or even fewer parameters, as a trade-off between accuracy and complexity. A detailed discussion about the accuracy of different models can be found in \citep{hughes_AccuracyFisheyeLens_2010}.

\subsection{Projection Distortions}

To get a full $360^\circ$ FOV, single-view wide FOV images are often projected onto the surface of a sphere \citep{sturm_GeneralImagingGeometry_2008}. But in practical applications, the image has to be "flattened" (rectified) before being displayed on the screen. The projection of a sphere onto a plane inevitably deforms the surface. Here we call such distortions generated in this projection process the projection distortions. Note that the sphere flattening process is similar to the map projection in cartography \citep{snyder_FlatteningEarthTwo_1993}. Indeed, the target surface does not have to be a plane, as long as it is developable. A developable surface means it can be unfolded or unrolled into a plane without distortion, such as a cylinder, cone, or plane. In computer vision tasks, specific attributes of structures or contents in the image may need to be preserved, e.g., shapes or distance, leading to many different kinds of projection methods. Based on the target developable surface, we can divide them into three categories, i.e., cylindrical projection, conic projection, and azimuthal projection. The main properties of these typical projections are summarized in Table \ref{tab:projection}.

In cylindrical projection, meridians are mapped to equally spaced vertical lines and circles of latitude are mapped to horizontal lines. There are two typical cylindrical projections, i.e., Mercator projection and Equirectangular projection. Specifically, the Mercator projection is a conformal projection, which preserves the angle and shape of objects. But the object size is inflated, which becomes infinite at the poles. The equirectangular projection maps meridians and circles of latitude with constant spacing (parallel lines of constant distance) while the shape of objects is not preserved. As one of the typical examples of conic projection, the equidistant conic projection can preserve the distances along the meridians proportionately. It is useful when the target region is along a latitude. Azimuthal projection maps the sphere surface directly to a plane, which includes three typical examples, i.e., gnomonic projection, orthographic projection, and stereographic projection, which are the specific cases in the unified camera model described in Section \ref{unifiedmodel} with $\xi = 0$, $\xi = \infty$ and $\xi = 1$, \citep{stevenson_NonparametricCorrectionDistortion_1996,jabar_PerceptualAnalysisPerspective_2017}, respectively. The stereographic projection here has the same geometric meaning as the stereographic distortion model in capturing rays-based methods.

\begin{table}[ht]
    \centering
    \caption{Properties of typical sphere projections \citep{snyder_FlatteningEarthTwo_1993,fenna_CartographicScienceCompendium_2007}.}
    \label{tab:projection}
    \begin{tabular}{ c l l}
    \toprule
    Name & Type & Property \\
    \midrule
    \multirow{2}{*}{Cylindrical} & Mercator  & Conformal \\
    \cmidrule{2-3}
    & Equirectangular & Equidistant \\ 
    \midrule
    Conic & Equidistant conic & Equidistant  \\
    \midrule
    \multirow{3}{*}{Azimuthal} & Gnomonic  & Great circles to lines \\
    \cmidrule{2-3}
    & Orthographic & Parallel projection \\
    \cmidrule{2-3}
    & Stereographic & Conformal \\
    \bottomrule
    \end{tabular}
\end{table}

\section{Image Rectification} \label{rectification}
As mentioned before, although wide-angle images have been widely used in many vision applications due to their large FOV, the perspective transformation assumed in a conventional pinhole camera is broken, resulting in object distortion in the wide-angle image. These geometrical distortions make it hard to estimate scene distance, area, and direction, and more importantly, they prevent all these images from feeding the off-the-shelf deep networks that are trained on normal images in this deep learning era. To address this issue, the first step in using wide-angle images is usually to correct them. Many such rectification methods have been introduced and improved since the wide FOV cameras emerged decades ago. We divide these methods into two groups, i.e., the traditional geometry-based methods and deep learning-based methods. In the former group, special points (especially vanishing points), straight lines, geometric shapes, or contents are taken as the regularization or guidance to rectify the distortion so that the rectified images can obey the perspective transformation again. In the latter group, parameters of the distortion model or the equivalent warp field that represents the transformation from the distorted image to the undistorted one are learned from large-scale training data, which is usually synthesized from normal images based on various wide FOV camera models or distortion models. In the following parts, representative methods of each group will be reviewed and discussed in detail.

\subsection{Geometry-based Methods}
Traditionally, image rectification is treated as an optimization problem where the objective function to be minimized can be some energy and/or loss terms that are used to measure the distortions in the image. For example, the straightness of lines is one of the most commonly used loss terms in most of the traditional methods. However, users may not only care about geometric lines but also some semantic content, such as faces in portraits or buildings in the scene. Accordingly, weight maps based on visual attention can be used in the objective function for a better perceptual result. Besides, when multi-view images are available, multi-view geometry constraints can also be leveraged to estimate accurate and robust distortion parameters. We present these methods as follows.

\subsubsection{Line-based Methods}
\begin{figure}[ht]
    \centering
    \includegraphics[width=0.5\textwidth]{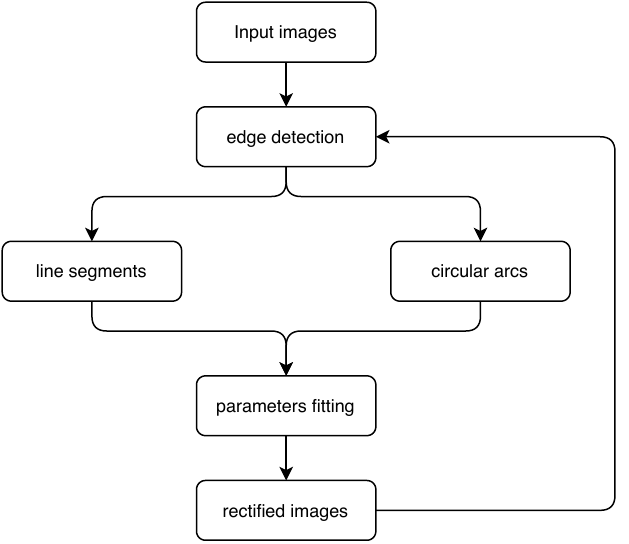}
    \caption{The workflow of line-based methods.}
    \label{fig:linebased}
\end{figure}

Among all structure information, straight lines are mostly leveraged as the regularization owing to the following reasons. First, they are intuitive and easy to understand. Second, they are sensitive to distortions caused by the wide FOV lens. Third, they can measure the distortion levels effectively and directly, e.g., based on the straightness of lines. As pointed out in \citep{zorin_CorrectionGeometricPerceptual_1995,devernay_AutomaticCalibrationRemoval_1995,devernay_StraightLinesHave_2001}, camera models follow perspective projection if and only if straight lines in the 3D world are still straight in the image. This is the golden rule in line-based rectification methods where the straightness of lines should be maximized or the curvature of line segments should be minimized in the rectification. The main framework of line-based methods is illustrated in Figure \ref{fig:linebased}. 

The first step of line-based methods is to detect lines in the distorted image, which itself is a non-trivial vision task. Usually, line detection is composed of two steps: edge detection, e.g., using the canny algorithm \citep{canny_ComputationalApproachEdge_1986}, and grouping of points on edges as line segments. When the distortion is small, line segments may be long enough to estimate parameters \citep{devernay_AutomaticCalibrationRemoval_1995,devernay_StraightLinesHave_2001}. However, when the distortion is large, a single line may break into too many small pieces, making parameters fitting unstable. Under this situation, these small pieces of segments should be merged into longer lines before fitting \citep{brauer-burchardt_AutomaticLensDistortion_2000, thormahlen_RobustLinebasedCalibration_2003}.

Once straight lines are collected in distorted image, parameters of distorted model can be estimated via non-linear optimization. It is known that every point $(x_u, y_u)$ on a 2D straight line satisfies:
\begin{equation}
    a x_u + b y_u + c = 0,
    \label{lineeq}
\end{equation}
where $a$, $b$, $c$ are scalar parameters that should be fitted for each line separately using all the points belonging to it. Undistorted coordinates $x_u$ and $y_u$ are mapped from the coordinates in distorted image via a mapping function $\mathcal{F}(\cdot)$, i.e., $x_u = \mathcal{F}_x(\mathbf{x_d};\mathbf{k}), y_u = \mathcal{F}_y(\mathbf{x_d};\mathbf{k})$, where $\mathbf{k}$ is the distortion parameter set and $\mathbf{x_d}=(x_d, y_d)$. Therefore, the sum square distance of all points to the lines is calculated as:
\begin{small} 
\begin{equation}
    L(\mathbf{\mathcal{X}_d}) = \sum_{i=1}^{K} \left( \sum_{j \in \mathbf{\Lambda_i}} \left( \frac{|a_i \mathcal{F}_x(\mathbf{x_d^{ij}}; \mathbf{k}) + b_i \mathcal{F}_y(\mathbf{x_d^{ij}};\mathbf{k)} + c_i|}{\sqrt{a_i^2 + b_i^2}} \right)^2 \right).
    \label{eq:line_loss}
\end{equation}
\end{small}
Here, $L(\cdot)$ denotes the loss function, $K$ is the number of lines in the image, $(a_i, b_i, c_i)$ is the fitted parameters for a specific line $l_i$, $\mathbf{\Lambda_i}$ is the index set of all points belonging to $l_i$, $\mathbf{\mathcal{X}_d} =\left \{ \left ( x_d^{ij},y_d^{ij} \right ) \mid i=1,\dots,K,j \in \mathbf{\Lambda_i} \right \}$. Intuitively, the distortion parameters can be estimated by minimizing this loss function. But in practice, it is hard to obtain accurate and robust estimation due to noise as well as the heavy computational cost arisen from nonlinear optimization. To address these issues, several other forms of loss function have been proposed, e.g., the slope of lines in \cite{ahmed_NonmetricCalibrationCamera_2001,ahmed_NonmetricCalibrationCamera_2005}, sum of residual error in   \cite{thormahlen_RobustLinebasedCalibration_2003}, and the sum of the angles between line segments belonging to the same line in \cite{kakani_AutomaticDistortionRectification_2020}.

If lines are detected via Hough transform-based methods, the most convenient way to represent a line is using the following equation \citep{prescott_LineBasedCorrectionRadial_1997}:
\begin{equation}
    x_u \cos{\theta} + y_u \sin{\theta} = \rho,
    \label{eq:hough_line}
\end{equation}
where $\rho$ is the perpendicular distance from the origin to the line and $\theta$ is the angle between the line and the horizontal axis. For each line $l_i$ in lines set with enough supporting points in the Hough space, we use $(\theta_i, \rho_i)$ to represent the line parameters. In the distorted image, the supporting points of a long straight line are broken into small groups because the line is detected as short pieces. After the image rectification, these short lines are connected as a long line that has the maximum number of support points. Thus, the loss function in Hough transform-based methods can be formulated as \citep{cucchiara_HoughTransformbasedMethod_2003}:
\begin{scriptsize}
\begin{equation}
    L(\mathbf{\mathcal{X}_d}) = \sum_{i=1}^{K} \left( \sum_{j \in \mathbf{\Lambda_i}} \left( \mathcal{F}_x(\mathbf{x_d^{ij}};\mathbf{k}) \cos{\theta_i} + \mathcal{F}_y(\mathbf{x_d^{ij}};\mathbf{k}) \sin{\theta_i} - \rho_i \right)^2 \right).
    \label{eq:loss_hough_line}
\end{equation}
\end{scriptsize}
Here, $K$, $\mathbf{\Lambda_i}$, $\mathbf{k}$, $\mathcal{F}_x(\cdot)$, and $\mathcal{F}_y(\cdot)$ have the same meaning as in Eq.~\eqref{eq:line_loss}.

In practice, due to noise and edge detection errors in the distorted image, $\theta_i$ and $\rho_i$ would not cluster into a point in the Hough space for a curved line. To make the Hough transform adapt the distorted line, distortion parameters are introduced into the hough space \citep{cucchiara_HoughTransformbasedMethod_2003,aleman-flores_WideAngleLensDistortion_2013,aleman-flores_AutomaticLensDistortion_2014,aleman-flores_LineDetectionImages_2014,santana-cedres_InvertibilityEstimationTwoParameter_2015}. The Hough transform that incorporates distortion parameters is called the extended Hough transform. Considering the computational efficiency and stabilization of the optimization, the dimension of the expended Hough space should not be too high. Therefore, only one parameter is introduced by choosing the one-parameter division model or one-parameter polynomial model as the distortion model \citep{cucchiara_HoughTransformbasedMethod_2003} in most cases. Furthermore, to make the estimation independent of the image resolution and avoid trivial small values, a proxy variable $p$ is estimated instead of $k$ in \cite{aleman-flores_WideAngleLensDistortion_2013,aleman-flores_LineDetectionImages_2014,aleman-flores_AutomaticLensDistortion_2014}. In later work, two distortion parameters are added in the extended Hough space via two-step optimization methods \citep{santana-cedres_InvertibilityEstimationTwoParameter_2015,santana-cedres_IterativeOptimizationAlgorithm_2016}. Although three or more parameters can be added similarly, it is not necessary to do so. Because the impact of high order coefficients decreases fast and the improvement becomes relatively small, while the computational cost and complexity increase quickly.

The above methods estimate the line parameters and distortion parameters based on linked small line segments, which are prone to noise and erroneous line detection. If the curved line in the distorted image could be detected directly instead of linking small pieces gradually, the estimation will be more accurate. As pointed out in \citep{brauer-burchardt_NewAlgorithmCorrect_2001,strand_CorrectingRadialDistortion_2005,wang_SimpleMethodRadial_2009,bukhari_RobustRadialDistortion_2010,bermudez-cameo_AutomaticLineExtraction_2015}, when one-parameter division model is taken to describe the distortion, the straight line in an undistorted image becomes a circular arc in the distorted image. If the distortion center is at the origin, points on lines in the undistorted image can be written as:
\begin{equation}
\left\{\begin{matrix}
    x_u = x_d / (1 + k r_d^2)
    \\ 
    y_u = y_d / (1 + k r_d^2)
    \end{matrix}\right.,
    \label{eq:oneparam_line_distortion}
\end{equation}
which are subjected to Eq.~\eqref{lineeq}:
\begin{equation}
    a \frac{x_d}{1+k r_d^2} + b \frac{y_d}{1+k r_d^2} + c = 0,
    \label{eq:lineeq_oneparam_form}
\end{equation}
i.e.,
\begin{equation}
    ck(x_d^2 + y_d^2) + a x_d + b y_d + c = 0.
\end{equation}
Here $k \neq 0$ is the distortion parameter. If the line does not pass through the origin, i.e., $c \neq 0$, then we have:
\begin{equation}
    x_d^2 + y_d^2 + \frac{a}{ck} x_d + \frac{b}{ck} y_d + \frac{1}{k} = 0. 
    \label{eq:lineeq_circle}
\end{equation}
This is a circle equation, implying that the straight line becomes a circle in the distorted image. More generally, if the distortion center is $(x_0, y_0)$, we have:
\begin{equation}
    (x_d - x_0)^2 + (y_d - y_0)^2 + \frac{a}{ck} (x_d - x_0) + \frac{b}{ck} (y_d - y_0) + \frac{1}{k} = 0,
    \label{eq:lineeq_circle_general}
\end{equation}
which can be denoted as:
\begin{equation}
    x_d^2 + y_d^2 + A x_d + B y_d + C = 0,
    \label{eq:lineeq_circle_general_param}
\end{equation}
\begin{align}
    A &= \frac{a}{ck} - 2 x_0,  \\ 
    B &= \frac{b}{ck} - 2 y_0, \\
    C &= x_0^2 + y_0^2 - \frac{a}{ck}x_0 - \frac{b}{ck}y_0 + \frac{1}{k}, \\
    0 &= x_0^2 + y_0^2 + A x_0 + B y_0 + C - \frac{1}{k}.
    \label{eq:lineeq_circle_general_params}
\end{align}
Given a group of points $(x_{d}, y_{d})$ on a curved line in the distorted image, the circle fitting algorithm \citep{bukhariAutomaticRadialDistortion2013,antunes_UnsupervisedVanishingPoint_2017} can be used to estimate $A, B, C$ in Eq.~\eqref{eq:lineeq_circle_general_param}. Moreover, given three arcs parameterized by $\{A_i, B_i, C_i, i=0,1,2\}$, $(x_0, y_0)$ can be calculated based on Eq.~\eqref{eq:lineeq_circle_general_params}, i.e.,
\begin{align*}
    \left\{
    \begin{array}{cc}
         (A_1 - A_0) x_0 + (B_1 - B_0) y_0 + (C_1 - C_0) = 0 &  \\
         (A_2 - A_1) x_0 + (B_2 - B_1) y_0 + (C_2 - C_1) = 0 & 
    \end{array}
    \right..
\end{align*}
The distortion parameter $k$ can be estimated using any of the three arcs' parameter and $(x_0, y_0)$ from Eq.~\eqref{eq:lineeq_circle_general_params}:
\begin{equation}
    \frac{1}{k} = x_0^2 + y_0^2 + A x_0 + B y_0 + C
    \label{eq:lineeq_circle_general_params_k}
\end{equation}

\begin{table*}[ht]
\scriptsize
    \centering
    \caption{Some of the typical constraints used in content-aware rectification methods.}
    \label{tab:contentaware}
  \newcommand{\tabincell}[2]{\begin{tabular}{@{}#1@{}}#2\end{tabular}}
    \begin{tabular}{ l c c c c c c c}
      \toprule
      \textbf{Method} & \textbf{Conformality} & \textbf{Points} & \textbf{Lines} & \textbf{Planes} & \textbf{Smoothness} & \textbf{Saliency} & \textbf{Boundary} \\
      \midrule
      \tabincell{c}{\cite{carroll_OptimizingContentpreservingProjections_2009} \\ \cite{sachtContentBasedProjectionsPanoramic2010}} & \checkmark & - & straightness & - & \checkmark & \tabincell{l}{standard deviation \\ face detection } & - \\
      \midrule
      \cite{kopf_LocallyAdaptedProjections_2009} & - & - & - & \checkmark & \checkmark & - & \checkmark \\
      \midrule
      \cite{carroll_ImageWarpsArtistic_2010} & - & \tabincell{l}{vanishing points \\ fixed points} & \tabincell{l}{straightness \\ orientation} & \checkmark & \checkmark & - & \checkmark \\
      \midrule
      \cite{sacht_ScalableMotionawarePanoramic_2011} & \checkmark & fixed points & straightness & - & \checkmark & - & - \\
      \midrule
      \cite{wei_FisheyeVideoCorrection_2012} & \checkmark & - & \tabincell{l}{straightness \\ orientation} & - & \checkmark & \tabincell{l}{standard deviation\\time variation\\motion saliency} & \checkmark \\
      \midrule
      \cite{kanamori_LocalOptimizationDistortions_2013} & - & - & straightness & - & - & - & - \\
      \midrule
      \cite{kim_AutomaticContentAwareProjection_2017} & - & - & straightness & - & - & \tabincell{l}{object detection\\motion saliency} & - \\
      \midrule
      \cite{jabar_ContentAwarePerspectiveProjection_2019} & - & - & \tabincell{l}{straightness\\orientation} & - & - & saliency detection & -\\
      \midrule
      \citep{shih_DistortionfreeWideanglePortraits_2019} & - & - & straightness & - & \checkmark & \tabincell{l}{segmentation\\face detection} & \checkmark \\
      \bottomrule
    \end{tabular}
\end{table*}

For images having large distortions, line detection or circle fitting is often prone to noise, e.g., unstable short line segments, curved lines in the 3D world, or wrong points near the lines. Usually, there are two ways to mitigate the issue. One is in an interactive way where straight lines are selected by humans \citep{carroll_OptimizingContentpreservingProjections_2009,carroll_ImageWarpsArtistic_2010,wei_FisheyeVideoCorrection_2012,kanamori_LocalOptimizationDistortions_2013}. The other way is to remove outliers and select the most informative lines iteratively. For example, lines with the most inner points are kept \citep{thormahlen_RobustLinebasedCalibration_2003} and the ones with inner points less than a threshold are removed \citep{kim_CorrectingRadialLens_2010}. Moreover, lower weights are assigned to lines near the distortion center because they are less informative than the ones far away. Similarly, lines that pass through the origin are deleted in \citep{benligiray_BlindRectificationRadial_2016}. \cite{zhang_RobustRadialDistortion_2015} selects good circular arcs regarding the histogram of the distortion parameters. Only the best three lines are selected for the estimation in \citep{zhang_LinebasedMultiLabelEnergy_2015}\footnote{http://cvrs.whu.edu.cn/projects/FIRC/}. \cite{antunes_UnsupervisedVanishingPoint_2017} leveraged Lines of Circle Centres (LCCs) for robust fitting.

In other work \citep{wildenauer_ClosedFormSolution_2013,jiang_EfficientLinebasedLens_2015}, the position of vanishing points is used as an extra constraint since all the parallel lines should pass through their vanishing points. Once we detect parallel lines, we can detect the vanishing points by calculating their intersections. Then, line parameters can be refined by leveraging the vanishing point constraint \citep{jiang_EfficientLinebasedLens_2015}. Besides, the sum of the distance from estimated vanishing points to the parallel lines can be used to measure the distortion \citep{yang_SimultaneouslyVanishingPoint_2016}, since vanishing points will scatter around the ground truth ones if there is distortion.

\subsubsection{Content-aware Methods}
When wide-angle images have large distortions \citep{carroll_OptimizingContentpreservingProjections_2009}, rectification using a single projection model may not preserve the straightness of lines and shapes of objects at the same time \citep{zorin_CorrectionGeometricPerceptual_1995}. Minimizing the overall distortions in a wide-angle image is to make some kind of trade-off between these two types of distortions. Since some contents in the image, e.g. the main building or human faces, are more important for a good perceptual result, they should be paid more attention during the rectification. These contents can be detected automatically or specified by users interactively, which usually contain salient semantic objects \citep{carroll_OptimizingContentpreservingProjections_2009,kopf_LocallyAdaptedProjections_2009,carroll_ImageWarpsArtistic_2010,sachtContentBasedProjectionsPanoramic2010,wei_FisheyeVideoCorrection_2012,kanamori_LocalOptimizationDistortions_2013}. In practice, many different kinds of constraints are often used together to construct the loss function for a better result. Some of them are listed in Table \ref{tab:contentaware}. These content-aware methods are trying to find a spatially varying warp field that transforms the distorted image to the corrected one while minimizing the distortion and keeping the pre-defined salient contents.

In interactive content-aware methods, users often specify points, lines, or regions that they care about. Therefore, the straightness and orientation of lines, e.g. the vertical lines of the building, are often used as constraints in the loss terms \citep{carroll_ImageWarpsArtistic_2010,wei_FisheyeVideoCorrection_2012,jabar_ContentAwarePerspectiveProjection_2019}. In \citep{kopf_LocallyAdaptedProjections_2009}, near-planar regions of interest can be annotated by users, whose planar attribute is kept in the rectified image via the deformation of the projection surface, e.g., a cylinder. In \citep{carroll_OptimizingContentpreservingProjections_2009}, the surface of a sphere is deformed to keep the user-specified constraints, e.g. horizontal lines to be horizontal and vertical lines to be vertical, after the image is corrected. In \citep{sachtContentBasedProjectionsPanoramic2010}, the loss function is constructed based on the constraints of the conformality of the mapping, the straightness of user-selected lines, and the smoothness of the warp field. They also leverage a saliency map to take the areas near line endpoints into account. Some other types of user-specified content are also used in \citep{carroll_ImageWarpsArtistic_2010}, e.g., vanishing point position and fixed points.

Recently, deep neural network-based methods have made significant progress in many computer vision tasks, including line detection and saliency map detection. Therefore, these two pre-processing steps in the above rectification methods can be accomplished by deep learning models. For example, \cite{kim_AutomaticContentAwareProjection_2017} extracts line segments using a deep model named Line Segment Detector (LSD) \citep{gromponevongioi_LSDLineSegment_2012} while \cite{jabar_ContentAwarePerspectiveProjection_2019} detects lines using EDLines proposed in \citep{akinlar_EDLinesRealtimeLine_2011}. Since these line detectors are trained on undistorted images, they may fail in spherical images where lines are curved. Therefore, the distorted image is usually rectified first by rectilinear projection such that lines are preserved and then the lines are detected by the line detectors. After that, points on the lines are projected back to the spherical coordinate and grouped, which are used to estimate the distortion parameters. In \citep{jabar_ContentAwarePerspectiveProjection_2019}, saliency map defined as the probability of object existence in the image \citep{kim_AutomaticContentAwareProjection_2017} is generated using ML-Net \citep{cornia_DeepMultilevelNetwork_2016}. In \citep{shih_DistortionfreeWideanglePortraits_2019}\footnote{https://github.com/Jason-xys/Wide-Angle-Portraits-Distortion-Correction}, the attention map is generated based on the union of the segmented human body and detected face.

\subsubsection{Multi-view Methods}
Image rectification heavily depends on the structure information in the image, e.g. straight lines. Compared to lines, points are more primitive features. Image with few lines may contain many distinctive keypoints. In this case, if multiple images of the same scene taken from different views are available, the image can be rectified based on point correspondence, as in the self-calibration methods \citep{faugeras_CameraSelfcalibrationTheory_1992, maybank_TheorySelfcalibrationMoving_1992,fraser_DigitalCameraSelfcalibration_1997,kang_CatadioptricSelfcalibration_2000}. And the other advantage that using points instead of lines is that the detection of points is faster, more stable and accurate than that of lines in distorted images. Specifically, when the camera is assumed to be a standard pinhole camera, point correspondence in multi-view images can be described by epipolar geometry \citep{hartley_MultipleViewGeometry_2003}. When images are distorted, the epipolar constraint will be broken \citep{zhang_EpipolarGeometryTwo_1996,stein_LensDistortionCalibration_1997,barreto_FundamentalMatrixCameras_2005}. Therefore, the deviation of corresponding points from the epipolar line can be used to measure the distortions. Minimizing the sum of the deviation distance leads to the best-fitted distortion parameters. Denoting that $[x_u, y_u]^\mathsf{T}$ and $[x_u^{\prime}, y_u^{\prime}]^\mathsf{T}$ are two correspondence points in two views without distortions, the epipolar line constraint is formulated as:
\begin{equation}
    [x_u, y_u, 1]^\mathsf{T} \mathbf{F} [x_u^{\prime}, y_u^{\prime}, 1] = 0, 
    \label{Fequation}
\end{equation}
where $\mathbf{F}$ is the $3 \times 3$ fundamental matrix \citep{hartley_MultipleViewGeometry_2003}. Assuming the images are taken by identical cameras and the distortion model in all views are the same, we have:
\begin{equation}
\left\{\begin{matrix}
     x_u = \mathcal{F}_x(\mathbf{x_d};\mathbf{k})
    \\ 
    y_u = \mathcal{F}_y(\mathbf{x_d};\mathbf{k})
    \end{matrix}\right.,
    \label{eq:multiview_distortion_mapping}
\end{equation}
where $\mathcal{F}_x$, $\mathcal{F}_y$, $\mathbf{x_d}$, and $k$ have the same meaning in Eq.~\eqref{eq:line_loss}. Substituting Eq.~\eqref{eq:multiview_distortion_mapping} into Eq.~\eqref{Fequation}, we can get the constraint for distorted points. If the points correspondence are known, the fundamental matrix and distortion parameters can be estimated by minimizing the loss function accordingly, i.e.,
\begin{equation}
    \min \limits_{\mathbf{F},\mathbf{k}} [\mathcal{F}_{xy}(\mathbf{x_d};\mathbf{k}), 1]^\mathsf{T} \mathbf{F} [\mathcal{F}_{xy}(\mathbf{x_d}^{\prime};\mathbf{k}), 1],
    \label{minf}
\end{equation}
In the standard pinhole camera model, the degree of freedom of $\mathbf{F}$ is eight. Therefore, if eight pairs of points are known, $\mathbf{F}$ can be estimated by solving a linear equation. Given more than eight pairs of points, it comes to the least-squares solution \citep{stein_LensDistortionCalibration_1997,pritts_MinimalSolversRectifying_2020}. When using wide FOV cameras, the optimization becomes complex since it also involves the distortion parameters. Although the distortion function with high order terms is also applicable in Eq.~\eqref{minf}, more parameters may not guarantee better results due to noise and unstable optimization \citep{hartley_ParameterfreeRadialDistortion_2005} while increasing computations.

If we reformulate $\mathbf{F}$ into the vectorization form, i.e., a nine-dimension vector $\mathbf{f}$, Eq.~\eqref{Fequation} can be rewritten as:
\begin{equation}
    [x_u x_u^{\prime}, x_u y_u^{\prime}, x_u, y_u x_u^{\prime}, y_u y_u^{\prime}, y_u, x_u^{\prime}, y_u^{\prime}, 1]^\mathsf{T} \mathbf{f} = 0. 
    \label{f9}
\end{equation}
When one-parameter distortion model is used, Eq.~\eqref{f9} can be formulated as a quadratic eigenvalue problem (QEP) \citep{fitzgibbon_SimultaneousLinearEstimation_2001, liu_CorrectingLargeLens_2014}, i.e.,
\begin{equation}
    (k^2 \mathbf{d}_1 + k \mathbf{d}_2 + \mathbf{d}_3)^\mathsf{T} \mathbf{f} = 0,
    \label{eq:qep}
\end{equation}
where $\mathbf{d}_1, \mathbf{d}_2, \mathbf{d}_3$ are vectors having the same size of $\mathbf{f}$, whose element is a function of $(x_u, y_u, x_u^{\prime}, y_u^{\prime}, k)$ \citep{liu_CorrectingLargeLens_2014}. The QEP can be solved using a quadratic eigenvalue solver given nine pairs of points to obtain the distortion parameter $k$ and fundamental matrix $\mathbf{F}$.

Note that $\mathbf{F}$ is a rank-2 matrix and $\det(\mathbf{F}) = 0$, which can be used as extra constraint to narrow the search space of the solution of Eq.~\eqref{minf} or Eq.~\eqref{eq:qep} \citep{Li05anon-iterative,kukelova_MinimalSolutionRadial_2011,liu_CorrectingLargeLens_2014}. Besides, if more views are available, the number of required point pairs can be reduced \citep{stein_LensDistortionCalibration_1997, steele_OverconstrainedLinearEstimation_2006}.

\subsection{Learning-based Methods}
To address the aforementioned demerits of traditional geometry-based methods, deep learning-based methods have been proposed in recent years. Given a distortion model, one simple idea is to learn its parameters from large-scale training data by regression. From another point of view, distorted images and rectified images can be seen as paired samples in two different domains, where each one can be transformed into the other via a warp field. Base on these two ideas, there are two main kinds of deep learning methods for image rectification, i.e., model-based methods that aim to predict the parameters of a specific distortion model and model-free methods that aim to learn the warp field or generate the rectified image. The most salient characteristic of model-free methods is that the distortion parameters are not involved in the framework and multiple distortion models can work together under one framework. Compared with traditional geometry-based methods, the target of model-based methods is the same as that of the two-stage methods, while the target of model-free methods is the same as that of the one-stage methods.

\begin{figure}[ht]
    \centering
    \includegraphics[width=0.5\textwidth]{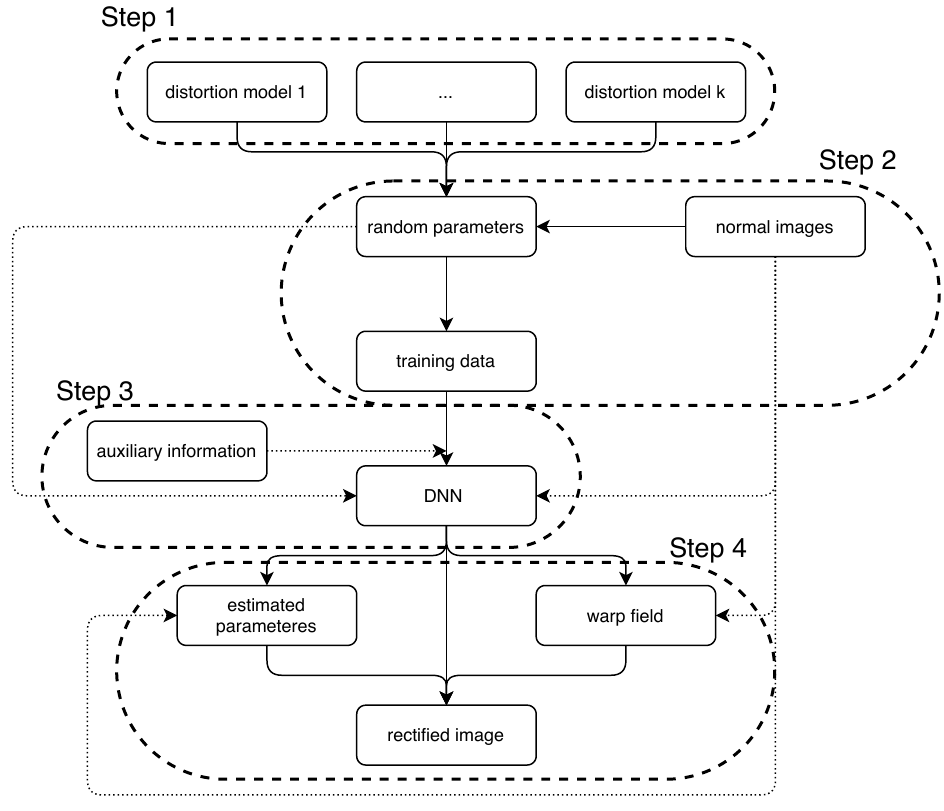}
    \caption{The general framework of the learning-based methods. In the first step, the candidate pool of distortion models is constructed. Then, one or more models can be chosen from it. Next, training pairs are generated by warping normal images according to the distortion model with randomly sampled parameters from a prior distribution. The sampled parameters and/or the generated warp field can be used as the ground truth. Then, a deep neural network is carefully designed as the key part of learning-based methods. Auxiliary information such as straight lines annotations and/or semantic segmentation maps can be incorporated to assist the training. Next, the distortion parameters or the warp field will be learned. Finally, the distorted image is rectified using the estimated parameters or warp field accordingly. In some cases, the model may learn parameters or warp field implicitly and output rectified image directly.}
    \label{fig:learning_flow}
\end{figure}

The most challenging issue for learning-based methods is their requirement for massive training data. Since it is hard to collect real-world paired training data, a typical solution is to generate synthetic training data based on distortion models. As described in Section \ref{distortionmodels}, there are many kinds of distortion models. So the first step is to choose a distortion model with parameters sampled from a prior distribution. The synthetic images paired with the original normal images are treated as training pairs. Besides, some semantics information and/or structure information can also be annotated, which can be leveraged to train a better model. Note that if the model is trained on a small dataset based on a specific distortion model, the generalizability will be limited. Therefore, a large-scale training dataset that covers as many distortions as possible is expected to train a useful model with good generalizability. The general framework of the learning-based methods is shown in Figure~\ref{fig:learning_flow}.

\subsubsection{Model-based Methods}
Model-based methods regress parameters of the explicit distortion model directly from the synthetic training data. For example, \cite{rong_RadialLensDistortion_2016} uses images from ImageNet \citep{deng_ImageNetLargescaleHierarchical_2009}\footnote{http://www.image-net.org/}

to synthesize distorted training images. Images with long lines are first selected and then the one-parameter division model is used to synthesize the distorted images. During training, it is formulated as a classification problem where the known distortion parameter $k$ is divided into 401 sub-classes. In the testing phase, a weighted average strategy is proposed to calculate the distortion parameter based on the predicted class probability. However, directly synthesizing distorted images from normal ones will generate black areas near image boundaries. To address this issue, \cite{bogdan_DeepCalibDeepLearning_2018}\footnote{https://github.com/alexvbogdan/DeepCalib} leverages the textcolor{red}{UCM model (refer to Section \ref{unifiedmodel})} to synthesize images by re-projecting images from a sphere. First, the panorama is projected onto the sphere surface under the unified camera model. Then, distorted images are generated via stereographic projection. Accordingly, the distortion parameters are the focal length $f$ and the distance $\xi$ from the projection center to the sphere center. Three different structures of networks are compared in the paper, including SingleNet, DualNet, and SeqNet.

\begin{table*}[ht]
\scriptsize
    \centering
    \caption{A summary of some typical model-based methods.}
    \label{tab:LP_table}
  \begin{threeparttable}[b]
  \newcommand{\tabincell}[2]{\begin{tabular}{@{}#1@{}}#2\end{tabular}}
    \begin{tabular}{ r c c c c c c}
      \toprule
      \textbf{Method} & \textbf{Size} & \textbf{Architecture} & \textbf{Model} & \textbf{Parameters} & \textbf{Information} & \textbf{Dataset}\\
      \cmidrule{1-7}
      \cite{rong_RadialLensDistortion_2016} & 256 & AlexNet & Division\tnote{3} & $k$ & - & ImageNet \citep{deng_ImageNetLargescaleHierarchical_2009}\\
      \cmidrule{1-7}
      \cite{yin_FishEyeRecNetMulticontextCollaborative_2018} & - & VGG & Poly.\tnote{4} & \tabincell{c}{$k_{1-5}$ \\ $m_u, m_v, u_0, v_0$} & Scene Parsing & ADE20K \citep{zhou_SemanticUnderstandingScenes_2019} \\
      \cmidrule{1-7}
      \cite{bogdan_DeepCalibDeepLearning_2018} & 299 & InceptionV3 & Stereographic\tnote{5} & $f, \xi$ & - & SUN360 \citep{xiao_RecognizingSceneViewpoint_2012}\tnote{1}{} \\
      \cmidrule{1-7}
      \cite{shi_RadialLensDistortion_2018} & 256 & \tabincell{c}{AlexNet\\ResNet18} & Division & $k$ & - & ImageNet \citep{deng_ImageNetLargescaleHierarchical_2009}\\
      \cmidrule{1-7}
      \cite{lopez_DeepSingleImage_2019} & 224 & DenseNet-161 & Poly. & $k_{1-2}, f, \theta, \phi$ & Horizon Line & SUN360 \citep{xiao_RecognizingSceneViewpoint_2012}\\
      \cmidrule{1-7}
      \tabincell{l}{\cite{xue_LearningCalibrateStraight_2019} \\ \cite{xue_FisheyeDistortionRectification_2020}} & 320 & \tabincell{c}{HG Net\\ResNet50} & Poly. & \tabincell{c}{$k_{1-5}$ \\ $m_u, m_v, u_0, v_0$} & Lines & \tabincell{c}{SUNCG \citep{song_SemanticSceneCompletion_2017}\\Wireframe \citep{huang_LearningParseWireframes_2018}}\\
      \cmidrule{1-7}
      \cite{liao_OIDCNetOmnidirectionalImage_2020} & - & \tabincell{c}{InceptionV3} & Poly. & \tabincell{c}{$k_0, k_2, k_3, k_4$ \\ $u_0, v_0$} & Prior Mask & \tabincell{c}{Oxford Building \citep{philbin_ObjectRetrievalLarge_2007}}\\
      \cmidrule{1-7}
      \cite{yang_UnsupervisedFisheyeImage_2020} & 128x128 & \tabincell{c}{WGAN\\VGG16} & Poly.  & $k_0, k_2, k_3, k_4$ & \tabincell{c}{Prior Mask\\Longest Line} & Place2 \citep{zhou_Places10Million_2018}\tnote{2} \\
      \bottomrule
    \end{tabular}
    
    \begin{tablenotes}
     \item[1] http://people.csail.mit.edu/jxiao/SUN360/
     \item[2] http://places2.csail.mit.edu/
     \item[3] The Typical Division Model in Table \ref{tab:fisheye-R}.
     \item[4] The Polynomial Radial Distortion Model in Table \ref{tab:fisheye-R}.
     \item[5] The Stereographic Model in Table \ref{tab:fisheye-theta}.
   \end{tablenotes}
    \end{threeparttable}
\end{table*}

Inspired by the traditional geometry-based methods that use straight lines in the images as distortion clues, some deep learning-based methods also explore these clues as guidance to get better results. For example, \cite{xue_LearningCalibrateStraight_2019,xue_FisheyeDistortionRectification_2020} proposes a new dataset, named the synthetic line-rich fisheye (SLF) dataset, which contains fisheye images, heatmaps of the distorted lines, rectified images, heatmaps of rectified lines, and the distortion parameters. These annotations are transferred from the wireframes dataset \citep{huang_LearningParseWireframes_2018}\footnote{https://github.com/huangkuns/wireframe} and SUNCG 3D dataset \citep{song_SemanticSceneCompletion_2017}\footnote{https://sscnet.cs.princeton.edu/}. A deep network composed of three cascade modules is utilized to do the rectification. The first module is used to detect distorted lines in the fisheye image. The second module takes the original fisheye image, the detected distorted lines, and their heatmap as inputs to predict the distortion parameters. The third module is a differentiable rectification layer, which aims to rectify the heatmap of distorted lines and distorted images given the predicted distortion parameters. In \citep{xue_FisheyeDistortionRectification_2020}\footnote{https://xuezhucun.github.io/LaRecNet/}, an attentive uncertainty regularization is introduced to add an attention mask to the $L1$ loss between the distorted image and the rectified image. Apart from the useful structure information of lines, semantic information is also explored for image rectification. For example, \cite{yin_FishEyeRecNetMulticontextCollaborative_2018} adds a scene parsing module to aid the rectification network. Specifically, the fisheye image first goes through a base network to obtain the encoded feature map. Then, a semantic segmentation head network is used to predict the semantic segmentation map from the encoded feature map. Finally, the distortion parameters can be estimated from a distortion parameter estimation head network, which takes the shallow feature maps of the base network, the encoded feature map, and the scene segmentation map as inputs. These feature maps can be seen as low-level, mid-level, and high-level information in the original image, which improves the prediction accuracy.

Most of the fisheye distortions have a fixed pattern, i.e., the two most popular types of fisheye distortion, barrel distortion and pincushion distortion, are radially symmetric and the distortion increases as the radius grows. Therefore, taking it as \textit{a priori} knowledge can help the network to converge faster and better. For example, \cite{shi_RadialLensDistortion_2018} proposes an inverted foveal layer specifically designed for barrel distortion, which can be inserted into the parameter regression network to accelerate the training process and obtain a smaller training and testing loss. \cite{liao_OIDCNetOmnidirectionalImage_2020} uses a prior attentive mask to help the parameter prediction. It is based on the following observations: 1) the distortion center is not far from the center of the lens, and 2) with the increase of the distance between the pixel and the distortion center, the distortion becomes larger. Specifically, in the first stage, a DC-Net takes the original image and a coarse mask map as inputs and predicts a refined mask map. Next, this refined mask and the original image are fed into a DP-Net to predict the distortion parameters. Multi-scale features in DP-Net are fused and the prediction is carried out in a cascaded way via sequential classification and regression to improve the accuracy. Some of the distortion parameters, e.g., tilt angle and focal length, are difficult to estimate since they are not directly observable in the image. To mitigate this issue, \cite{lopez_DeepSingleImage_2019} uses proxy variables instead of the extrinsic and intrinsic parameters, which have close relationships to visual clues and can be estimated easily.

Comparisons of the typical model-based methods are summarized in Table \ref{tab:LP_table}. It can be seen that 1) the polynomial model and the one-parameter division model are most commonly used; 2) usually less than five orders are used in the polynomial model; 3) the size of the training image is small, e.g., $256 \times 256$, and 4) each model is trained using different synthetic datasets. The synthetic datasets are different from each other in two aspects, the standard image datasets and the distortion models. Some of them use different standard image datasets, or use the same image dataset but with different distortion models, or are different in both.

\subsubsection{Model-free Methods}
Since model-based methods aim to estimate the parameters corresponding to a specific distortion model, it is inflexible for them to adapt to various distortion models in one framework. By contrast, model-free methods do not estimate the distortion parameters but try to learn the warp field that transforms the distorted image to the undistorted one by per-pixel displacement vector. Because the warp field does not bind with the distortion model, it is possible to represent multiple types of distortion models in one warp field, leading to a promising general solution. \cite{li_BlindGeometricDistortion_2019}\footnote{https://github.com/xiaoyu258/GeoProj} pre-defines six different types of distortion models and designs two types of networks, i.e., GeoNetS and GeoNetM, to predict the warp field. GeoNetS estimates a single-model distortion field, which is calculated via the predicted distortion parameters and supervised explicitly by the ground truth flow field. However, GeoNetS is limited to only one specific distortion model once being trained. To estimate the distortion field that covers all six types of models using one network, GeoNetM is proposed which has a multi-task structure, one head for classification of the distortion types and the other head for estimation of the flow field. It uses the estimated flow field to fit the parameters of the predicted type of distortion model. Finally, the flow field is regenerated based on the distortion model with the fitted parameters, which can be seen as a fusion of both tasks, making the result more accurate. \cite{liao_ModelFreeDistortionRectification_2020} also proposes a model-free learning framework that can handle multiple types of distortion models in one deep model and expects better generalizability. Specifically, 16 distortion models are leveraged to synthesize the training data. Instead of estimating the heterogeneous distortion parameters, they propose estimating the distortion distribution map (DDM), which could cover any distortion model in the same form. DDM describes distortion as the ratio between the coordinates of the same pixel in the distorted and rectified image, rather than the movement or displacement of the pixel. They use an encoder-decoder network to estimate DDM, which guides the extraction of semantic information from the distorted image. Meanwhile, they also use another encoder to learn structure features from the canny edge map. Next, the semantic information and structure features are fused via an attention map. Finally, a decoder uses the fused feature to predict the rectified image.

\begin{table*}[ht]
  \centering
  \caption{A summary of some typical model-free learning methods.}
  \label{tab:LF_table}
  \begin{threeparttable}[b]
  \newcommand{\tabincell}[2]{\begin{tabular}{@{}#1@{}}#2\end{tabular}}
    \begin{tabular}{l c c c c c}

      \toprule
      \textbf{Method} & \textbf{Size} & \textbf{Architecture} & \textbf{Model} & \textbf{Information} & \textbf{Dataset}\\
      \cmidrule{1-6}
      \cite{li_BlindGeometricDistortion_2019} & - & Encoder-Decoder & 6 models & - & -\\
      \cmidrule{1-6}
      \cite{lorincz_SingleViewDistortion_2019} & 640x192 & ResNet18 & TPS pairs & Scene Parsing & \tabincell{c}{KITTI \citep{geiger_AreWeReady_2012}\\Carla \citep{pmlr-v78-dosovitskiy17a}} \\
      \cmidrule{1-6}
      \cite{liao_ModelFreeDistortionRectification_2020} & 256x256 & Encoder-Decoder & 16 models & Canny Edges & MS-COCO \citep{lin_MicrosoftCOCOCommon_2014}\tnote{1} \\
      \cmidrule{1-6}
      \cite{liao_DRGANAutomaticRadial_2020} & 256x256 & \tabincell{c}{U-Net\\VGG19} & Poly. & - & MS-COCO \citep{lin_MicrosoftCOCOCommon_2014} \\
      \bottomrule
    \end{tabular}
    \begin{tablenotes}
     \item[1] https://cocodataset.org/\#home
  \end{tablenotes}
  \end{threeparttable}
\end{table*}

In some other works, the rectification problem is treated as an image-to-image translation problem, where the standard perspective image and the distorted image are seen as samples from two different domains. As a powerful tool in domain transformation, GAN \citep{NIPS2014_5423} is also introduced in image rectification. For example, \cite{liao_DRGANAutomaticRadial_2020} presents the DR-GAN, a conditional generative adversarial network for automatic radial distortion rectification. The rectified image is generated by the generator and a low-to-high perceptual loss is used to improve the output image quality. Further on, \cite{yang_UnsupervisedFisheyeImage_2020} adds a prior attentive map as in \citep{liao_ModelFreeDistortionRectification_2020} and takes the longest straight line in the standard image as the weighting map when calculating forward and backward loss. The attentive map is used to quantify the distortion spatially. Forward loss is defined in the distorted image domain while backward loss is defined in the standard image domain. A summary of the model-free methods is presented in Table \ref{tab:LF_table}.

\subsection{Discussion}
\subsubsection{Traditinal Geometry-based Methods}

Traditional geometry-based methods can be divided into two categories, i.e., the one-stage methods and the two-stage methods. In the former category, a warp field is optimized directly, e.g., \citep{kopf_LocallyAdaptedProjections_2009,carroll_OptimizingContentpreservingProjections_2009,carroll_ImageWarpsArtistic_2010}. These methods are usually carried out in an interactive manner and leverage the constraint of user-specified contents that need to be preserved or adjusted. However, it may be challenging for a user without domain knowledge to select the proper contents that lead to a satisfying result.

In two-stage methods, some preliminary tasks like line detection, circle fitting, vanishing points localization, or point correspondence in multi-view images are first carried out automatically. And then distortion parameters are estimated based on the constraints of these elements. Although every step of the procedure can be completely automatic, the errors in each stage will accumulate to the deterioration of the estimation of the parameters. More importantly, the two procedures are always coupled together, making them hard to be disentangled and optimized separately. To address this issue, some iterative algorithms are proposed to refine the estimate in a loop, but the errors may accumulate step by step in the iterative pipeline.

Generally, on the one hand, the demerit of traditional geometry-based methods is the high complexity that too many hyper-parameters in each sub-task need to be tuned carefully, e.g., the thresholds in edge detection and line segments grouping. Besides, the empirically selected parameters may not work well in various scenarios in practical applications. On the other hand, the merit of traditional geometry-based methods is that the solutions are always analytical and explainable where the outputs of each step have explicit meanings. Besides, they have good generalizability to different sizes of images.

\subsubsection{Learning-based Methods}

Similar to the deep learning methods for other computer vision tasks, the performance of learning-based methods for image rectification also depends on large-scale training data. To our best knowledge, there is no real-world paired training data available for image rectification. Existing methods always use their own training data synthesized based on specific types of distortion models, making it difficult to compare their performance.

Moreover, the synthetic training data depends on the chosen distortion models, which consequently limits the generalizability of the trained model. Although one can sample various parameters from a prior distribution to generate large amounts of training data, they are limited to the exact specific type of distortion model. Even if multiple distortion models can be used to build the training dataset \citep{li_BlindGeometricDistortion_2019,liao_ModelFreeDistortionRectification_2020}, there is still a gap between the synthetic training images and real-world wide-angle images from various wide FOV cameras like fisheye cameras and omnidirectional cameras. More efforts should be made to bridge the gap.

Besides, leveraging supervision from mid-level guidance like lines and high-level guidance like semantics has attracted increasing attention in recent years. For example, when mid-level guidance like straight lines is used as extra supervision \citep{xue_FisheyeDistortionRectification_2020}, the network is supervised to learn how the structural elements (e.g., lines) in the image should be rectified, thereby obtaining better generalizability. One step further, high-level guidance can provide more abundant information about regions than the mid-level ones, e.g., scene parsing. It has been proven that incorporating high-level guidance into the network can improve the rectification result \citep{yin_FishEyeRecNetMulticontextCollaborative_2018,lorincz_SingleViewDistortion_2019}. Therefore, it is promising and worth trying to explore other forms of high-level guidance, e.g., depth or instance segmentation.

\subsubsection{Relationship}

Generally, no matter the traditional geometry-based methods or the learning-based methods, they both try to estimate the mapping between the distorted image and the undistorted one. The distortion model, as a bridge between the two domains, plays an important role in this estimation. In most cases, the estimation is equivalent to predicting the parameters of the distortion model. In traditional geometry-based methods, the prediction is formulated as an optimization problem, where the loss function that measures the distortion is minimized. In learning-based methods, it is usually formulated as a regression problem, where the parameters are regressed by minimizing the distance between the predicted parameters and the ground truth ones.

Existing learning-based methods depend on the distortion model implicitly or explicitly, whereas traditional geometry-based methods can get rid of distortion models completely, e.g., \citep{sachtContentBasedProjectionsPanoramic2010}. Although distortion parameters may not be needed as supervision in model-free learning methods, the distortion model has to be assumed as \textit{a prior} knowledge for synthesizing the training data. Therefore, the distorted image or the equivalent supervision signal, i.e., warp field \citep{liao_ModelFreeDistortionRectification_2020}, are still based on the distortion model. By contrast, traditional geometry-based methods use straight lines as constraints by maximizing their straightness, which is distortion model agnostic.

Traditional geometry-based methods can only assume one distortion model in one solution since different models could lead to different forms of loss functions which are very hard to formulate in a single framework. For example, the same distortion can be generated by different models with different parameters. By contrast, owing to the strong representation capacity of deep networks, learning-based methods (or model-free methods specifically) could adapt to multiple distortion models in one solution \citep{li_BlindGeometricDistortion_2019,liao_ModelFreeDistortionRectification_2020}, as long as adequate training data that covers these distortion models are provided. However, even with multiple distortion models, learning-based methods commonly have the problem of generalization, i.e., the deep model that is trained on one dataset may perform poorly on another dataset, or the one trained on the synthetic dataset can not work well on real images. In contrast to that, geometry-based methods are invariant to the domain of the images. No matter if the image is synthetic or real, or from an unknown dataset, they can provide consistent and explainable outputs, which is very difficult for learning-based methods.

Learning-based methods usually run faster than traditional geometry-based methods, especially the ones including an iterative refinement process, since learning-based methods only need one forward-pass computation in the testing phase, which can be accelerated by using modern GPUs. However, when the size of the input image increases, the computation cost of learning-based methods will increase accordingly, while that of traditional methods may almost stay the same. Because the number of lines will not increase as the size of the image increases. Note that the computation of edge detection is relatively small compared with the optimization procedure. Compared with learning-based methods, geometry-based methods have to decide many hyper-parameters empirically and some of them are vital to the performance. In contrast, learning-based methods not only have fewer hyper-parameters but also are not that sensitive to them.

\section{Performance Evaluation} \label{performance}
The evaluation of the image rectification can be carried out both qualitatively by subjective visual comparison and quantitatively according to objective metrics. However, as far as we knew, the lack of a benchmark dataset makes the evaluation difficult, which should be an important future work as will be discussed in Section~\ref{future}. In this paper, we provided an evaluation of typical rectification methods by collecting and analyzing the results from different methods in the existing literature. We also provide a strong baseline model and carry out an empirical study of different distortion models on synthetic datasets and real-world wide-angle images.

\subsection{Experiment Settings}
\begin{figure*}[ht]
    \centering
    \includegraphics[width=1.\textwidth]{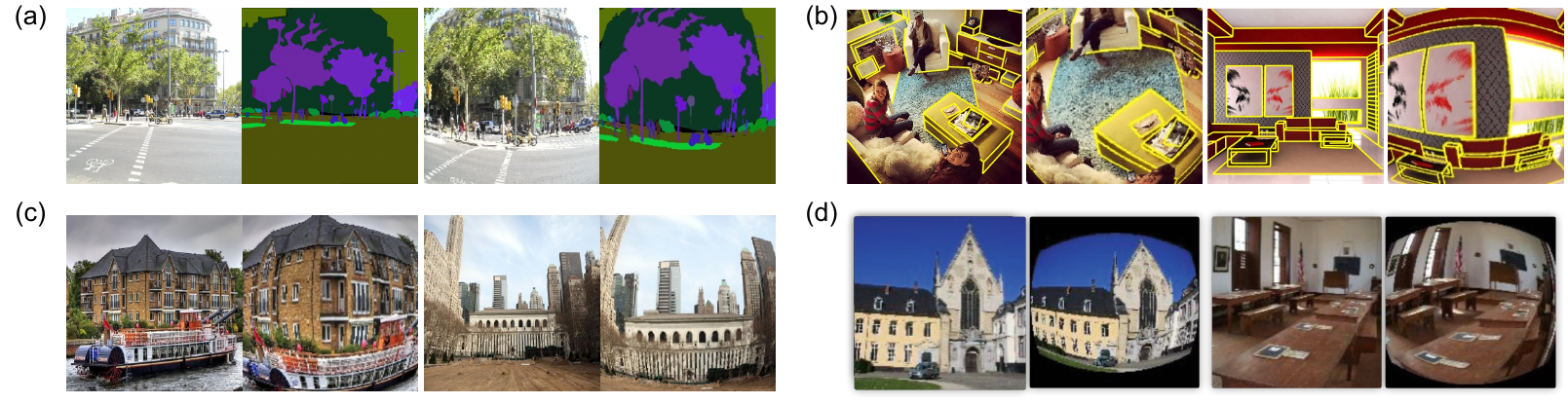}
    \caption{Some examples from different synthesized training sets. (a) Distorted images with scene parsing annotations synthesized from ADE20k \citep{zhou_SemanticUnderstandingScenes_2019}. The figure is reproduced from \citep{yin_FishEyeRecNetMulticontextCollaborative_2018}. (b) Distorted images with wireframe annotations synthesized from WireFrame dataset \citep{huang_LearningParseWireframes_2018}. The figure is reproduced from \citep{xue_FisheyeDistortionRectification_2020}. (c) Synthetic images from PLACE2 dataset \citep{zhou_Places10Million_2018}. The figure is reproduced from \citep{yang_UnsupervisedFisheyeImage_2020}. (d) Distorted images synthesized from ImageNet \citep{deng_ImageNetLargescaleHierarchical_2009}. The figure is reproduced from \citep{rong_RadialLensDistortion_2016}.}
    \label{fig:dataset}
\end{figure*}

\textbf{Datasets.} The ability of the learning-based methods partly depends on the training set. In early work, e.g., \citep{rong_RadialLensDistortion_2016}, the distorted training data is synthesized directly from standard images from the ImageNet dataset \citep{deng_ImageNetLargescaleHierarchical_2009} using randomly chosen distortion parameters. Since extra semantic or structure information has been proved useful for facilitating the training of deep networks, some datasets in related high-level computer vision tasks are also utilized for synthesizing distortion training images. For example, in \citep{yin_FishEyeRecNetMulticontextCollaborative_2018}, scene parsing annotations in ADE20k dataset \citep{zhou_SemanticUnderstandingScenes_2019}\footnote{https://groups.csail.mit.edu/vision/datasets/ADE20K/}
can be used as the high-level semantic supervision. Similarly, the network in \citep{lorincz_SingleViewDistortion_2019} is supervised by extra semantic labels from KITTI odometry dataset \citep{geiger_AreWeReady_2012}\footnote{http://www.cvlibs.net/datasets/kitti/}
and synthetic images via the Carla driving simulator \citep{pmlr-v78-dosovitskiy17a}\footnote{https://carla.org/}. Lines, as the most common structure, are used as the extra supervision in several works \citep{lopez_DeepSingleImage_2019,xue_LearningCalibrateStraight_2019,liao_ModelFreeDistortionRectification_2020,yang_UnsupervisedFisheyeImage_2020}. For example, the proposed line-rich dataset in \citep{xue_LearningCalibrateStraight_2019,xue_FisheyeDistortionRectification_2020} provide both the ground truth distortion parameters and the 2D/3D line segment annotations in man-made environments. Consequently, the LaRecNet trained on this dataset achieved state-of-the-art results. Some samples of the synthetic datasets are shown in Figure \ref{fig:dataset}.

\textbf{Metrics.} When the ground truth image is known, the difference between the rectified image and the original image can measure the accuracy of the rectification. PSNR and SSIM \citep{wang_ImageQualityAssessment_2004} are two widely used image quality assessment metrics with known reference, where the former accounts for mean square error and the latter accounts for the structural difference. We adopted them as objective evaluation metrics in this paper. Besides, in order to precisely measure the geometric accuracy of the rectified image, some new metrics have been proposed in  \citep{rong_RadialLensDistortion_2016,xue_FisheyeDistortionRectification_2020}. For example, precision and recall can be used to measure if the pixels on the distorted lines are still on the straight lines after rectification \citep{xue_FisheyeDistortionRectification_2020}. The precision and recall are calculated by
\begin{equation}
    Precision = |P \cap G| / |P|,\\
    Recall = |P \cap G| / |G|,
    \label{eq:precision}
\end{equation}
where $P$ is the set of pixels on lines in the rectified image and $G$ is the set of line pixels in the ground truth (standard) image. $|\cdot|$ denotes the number of pixels in the set. $|P \cap G|$ is the number of correctly rectified pixels (positive samples). Precision measures the ratio of correctly rectified line pixels in the rectified image and recall measures the ratio of the line pixels in the ground truth image that are correctly rectified. The overall performance is measured by the maximal F-score of every pair of precision and recall at different thresholds \citep{xue_FisheyeDistortionRectification_2020}. The F-score is defined as:
\begin{equation}
    F = \frac{2 \cdot Precision \cdot Recall}{Precision + Recall}.
    \label{eq:Fscore}
\end{equation}

Furthermore, the accuracy of the estimated distortion parameters can be measured by the reprojection error (RPE)  \citep{xue_FisheyeDistortionRectification_2020}. Given the ground truth and the estimated distortion parameters, every pixel on the distorted image can be re-projected to the rectified image using the inverted distortion model. If the estimated parameters are accurate, the distance between the re-projected pixels and the ground truth should be zero.

\begin{table*}[ht]
    \centering
    \caption{PSNR/SSIM of state-of-the-art methods on different test sets.}
    \label{tab:PSNR}
    
  \begin{threeparttable}
    \newcommand{\tabincell}[2]{\begin{tabular}{@{}#1@{}}#2\end{tabular}}
    \begin{tabular}{ r  c  c c c c}
      \toprule
      {} & {\textbf{Model}} & \textbf{SLF} & \textbf{FV} & \textbf{FR} & \textbf{DR} \\

      \hline
      \cite{bukhariAutomaticRadialDistortion2013} & {-} & 9.34/0.18&9.84/0.16&11.47/0.2429&12.52/0.3082   \\
      \cite{aleman-flores_AutomaticLensDistortion_2014} & {-} & 10.23/0.26&10.72/0.30&-/-&13.22/0.3311  \\
      \hline
      \cite{rong_RadialLensDistortion_2016} & {DM} & 12.92/0.32&11.81/0.30&13.08/0.3356&13.96/0.3741   \\
      \cite{yin_FishEyeRecNetMulticontextCollaborative_2018} & {Poly.} & -/-&-/-&14.96/0.4129&-/-   \\
      \cite{xue_FisheyeDistortionRectification_2020} & {Poly.} & \textbf{28.06/0.90}&\textbf{22.34/0.82}&-/-&-/-    \\
      \cite{liao_DRGANAutomaticRadial_2020} & {-} & -/- & -/- & -/- & \textbf{16.59/0.6835}   \\
      \hline
      Our baseline & {DM} & 25.10/0.83 & -/- & \textbf{24.76/0.81} & -/-  \\
      \bottomrule
    \end{tabular}
    
  \end{threeparttable}
\end{table*}

\subsection{Performance Evaluation of State-of-the-art Methods} 
We collected the reported results in the state-of-the-art (SOTA) works and analyzed the performance accordingly. Here we selected six SOTA methods for comparison, among which \cite{bukhariAutomaticRadialDistortion2013} and \cite{aleman-flores_AutomaticLensDistortion_2014} are two representative traditional geometry-based methods, \cite{rong_RadialLensDistortion_2016} is a typical and seminal model-based method, \cite{liao_DRGANAutomaticRadial_2020} is a model-free method. In order to verify the effectiveness of extra information and guidance in parameter regression, we also included \cite{yin_FishEyeRecNetMulticontextCollaborative_2018} and \cite{xue_FisheyeDistortionRectification_2020} in the evaluation. We chose PSNR, SSIM, F-score, and RPE as four objective metrics for the evaluation. Four datasets were used by referring to \cite{yin_FishEyeRecNetMulticontextCollaborative_2018},\cite{xue_FisheyeDistortionRectification_2020}, and \cite{liao_DRGANAutomaticRadial_2020}, named as SLF (the \textbf{S}ynthetic \textbf{L}ine-rich \textbf{F}isheye test set used in \citep{xue_FisheyeDistortionRectification_2020}), FV (the \textbf{F}isheye \textbf{V}ideo test set used in \citep{xue_FisheyeDistortionRectification_2020}), FR (the test set used in \textbf{F}ishEye\textbf{R}ecNet \citep{yin_FishEyeRecNetMulticontextCollaborative_2018}), and DR (the test set used in \textbf{DR}-GAN \citep{liao_DRGANAutomaticRadial_2020}), respectively. The DR dataset was synthesized using the even-order polynomial distortion model with six distortion parameters. By contrast, nine distortion parameters were used in the generation of SLF and FR datasets. Furthermore, DR dataset consists of 30,000 training image pairs while SLF dataset has 46,000 training samples and FR dataset contains 24,500 samples. Considering the distortion model and the number of training samples, SLF dataset is more complex than DR and FR dataset. FV dataset contains both synthetic images and real fisheye videos, thereby it can be used to test the generalizability of the methods.

The PSNR and SSIM of six SOTA methods tested on these four datasets were summarized in Table \ref{tab:PSNR}. Although the scores of some methods are not available, the overall trends still make sense. First of all, deep learning-based methods have higher PSNR and SSIM scores than traditional methods. As a pioneer work, \cite{rong_RadialLensDistortion_2016} divides the range of the parameters into 401 classes, so the estimation accuracy is limited. But owing to the powerful capacity of the deep neural network, it still obtains a gain of 1dB $\sim$ 3dB PSNR over traditional methods. Second, extra semantic annotations can benefit rectification. \cite{yin_FishEyeRecNetMulticontextCollaborative_2018} predicts the rectified image and the scene parsing result simultaneously. With the extra guidance of the semantics and the direct L2 loss on the ground truth image, it improves the PSNR by nearly 2dB compared to \citep{rong_RadialLensDistortion_2016}. Similarly, \cite{xue_FisheyeDistortionRectification_2020} uses the straight line annotations to guide the rectification like the traditional methods do. The supervision of the straight lines significantly advances the learning methods, i.e., leading to an overall 10dB improvement. From Table \ref{tab:RPE}, we can also see that points on lines are projected back to the straight lines after rectification. The principle that lines should be straight after rectification is also useful in learning-based methods.

In most of the learning-based methods, L1 or L2 loss is used between the corrected image and the ground truth, but these two losses can not handle details well \citep{isola_ImagetoImageTranslationConditional_2017}. In \citep{rong_RadialLensDistortion_2016}, perception loss is introduces into image rectification. Without using extra annotations, it leverages perception loss to supervise the training and achieves comparable results with \citep{yin_FishEyeRecNetMulticontextCollaborative_2018}. Like other deep learning methods, the fusion of multi-scale and multi-stage features is also useful in image rectification. In \citep{xue_FisheyeDistortionRectification_2020}, both global and local features are used to predict the distortion parameters, and then the average value is taken as the final result. \cite{rong_RadialLensDistortion_2016} adopt a U-Net \citep{ronneberger_UNetConvolutionalNetworks_2015} like architecture and use multi-level features to generate the corrected image.

\begin{table}[ht]
\scriptsize
  \centering
  \caption{F-score and RPE of state-of-the-art methods.}
    \label{tab:RPE}
    
  \begin{threeparttable}
    \begin{tabular}{ r c c | c c }
      \toprule
      {} & \multicolumn{2}{c|}{\textbf{F-score}} & \multicolumn{2}{c}{\textbf{RPE}} \\
      \cmidrule{2-5}
      {} & SLF & FV & SLF & FV \\
      \cmidrule{2-5}
      \cite{bukhariAutomaticRadialDistortion2013} & 0.29&- & 164.75&156.3 \\
      \cite{aleman-flores_AutomaticLensDistortion_2014} & 0.30&- & 125.42&125.31 \\
      \hline
      \cite{rong_RadialLensDistortion_2016} & 0.33&- & 121.69&125.31 \\
      \cite{xue_FisheyeDistortionRectification_2020} & \textbf{0.82}&- & \textbf{0.33}&\textbf{1.68} \\
      \bottomrule
    \end{tabular}
    
  \end{threeparttable}
\end{table}

\subsection{A Strong Baseline and Benchmark}
\textbf{Dataset\footnote{The source code, dataset, models, and more results will be released at: \url{https://github.com/loong8888/WAIR}.}.} Existing methods generate their training samples using standard images from different datasets based on different distortion models or with parameters sampled from different distributions. Consequently, it is hard to compare all the methods in the same setting. To mitigate this issue, we built a benchmark by synthesizing training and test images from three source datasets, ADE20k dataset \citep{zhou_SemanticUnderstandingScenes_2019}, WireFrame dataset \citep{huang_LearningParseWireframes_2018} and COCO dataset \citep{lin_MicrosoftCOCOCommon_2014} based on three distortion models, i.e., the \textbf{F}ield-\textbf{O}f-\textbf{V}iew distortion model in Row 4 of Table 2 (denoting ``FOV''), the typical \textbf{D}ivision \textbf{M}odel in Row 5 of Table 2 (denoting ``DM''), and the \textbf{E}quidistant \textbf{D}istortion model in Table 3 (denoting ``ED''). Details about these distortion models can be found in Section \ref{distortionmodels}. The three datasets are the most common ones in the vision community, and are also often used in rectification, while the three distortion models are the most simple and typical ones for wide FOV cameras.

The ADE20K dataset contains 20k images for training and 2k images for testing, while the WireFrame dataset contains 5k images for training and 462 images for testing. As for the COCO dataset, we used the 40k images in the test set to generate the training samples and the 5k images in the validation set to generate the test samples. Each original image was center-cropped with maximum size at the height or width side, which was then resized to $257 \times 257$. The distortion parameter of each distortion model is sampled from a uniform distribution within a pre-defined range, i.e., $[-0.02, -1]$ for the one-parameter division model, $[0.2, 1.2]$ for the FOV model, and $[0.7, 2]$ for the equidistant model. The training samples are synthesized on the fly during training.

\begin{figure}[ht]
    \centering
    \includegraphics[width=1.0\linewidth]{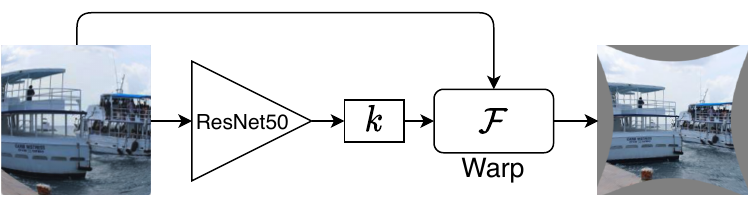}   
    \caption{The diagram of the proposed baseline model.}
    \label{fig:baseline_nn}
\end{figure}

\textbf{Network Architecture.} We used ResNet50 \citep{he_DeepResidualLearning_2016} pre-trained on ImageNet\citep{deng_ImageNetLargescaleHierarchical_2009} as the backbone network which is a widely used and \textit{de facto} standard structure in deep learning community., and changed the output channel of the FC layer to one for predicting the distortion parameter $k$. We devised a differentiable warp module to embody the warp function $\mathcal{F}$, which takes the distorted image and the estimated distortion parameter $k$ as inputs and outputs the rectified image. L1 loss between the rectified image and the ground truth image was minimized during the training. The whole network was trained end-to-end. The structure of the proposed baseline model is illustrated in Figure~\ref{fig:baseline_nn}.

\textbf{Results.} We trained three baseline models for each of the distortion models separately on the corresponding training dataset synthesized from the ADE20k dataset. Specifically, for each source dataset, we synthesized training and test samples using one of the distortion models, respectively. Therefore, we built nine datasets in total, three for each source dataset. Each model was only trained on the corresponding synthetic ADE20k training set but tested on all the synthesized test sets. For simplicity, we used the names of the distortion models and the original datasets to denote the synthetic datasets and the corresponding deep neural network models that were trained on them. For example, the FOV ADE20k dataset denotes the synthetic dataset generated from the ADE20k dataset based on the FOV distortion model, while the FOV ADE20k model denotes the deep model trained on the FOV ADE20k dataset. The PSNR and SSIM of the test results were summarized in Table \ref{tab:baseline_psnr}. The scores in each row are the results of one specific deep model tested on all the test sets, while the scores in each column are the results of different deep models tested on the same test set.

\begin{figure}[ht]
    \centering
    \includegraphics[width=1\linewidth]{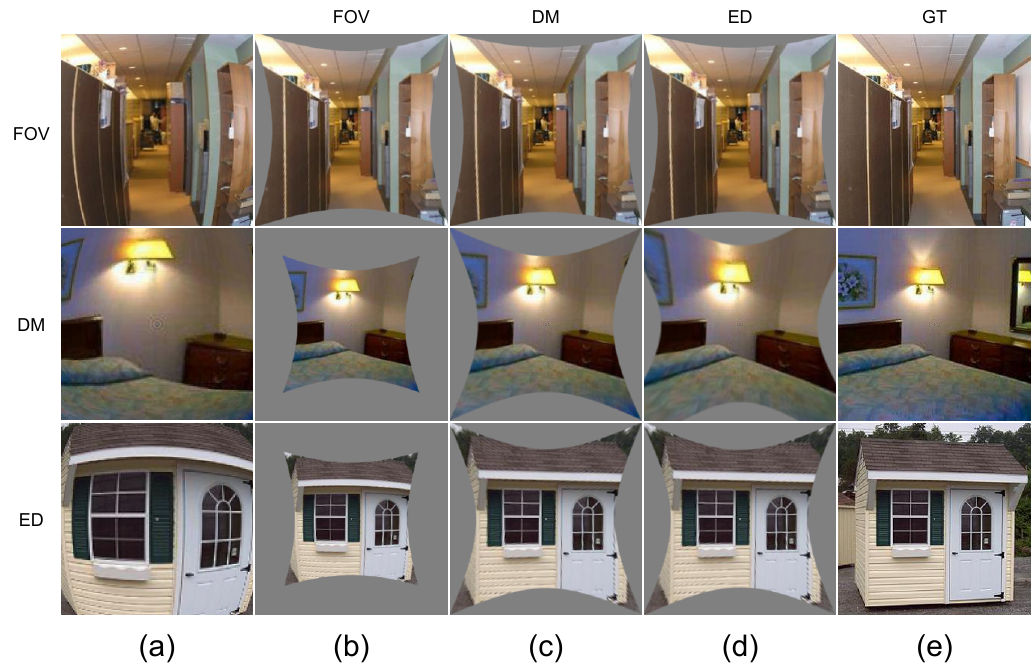}
    \caption{Results of our baseline models on synthetic images. (a) The synthetic images using the three distortion models. The original images are from the ADE20k test set. (b)-(f) The rectified results of the FOV ADE20k model, the DM ADE20k model, and the ED ADE20k model. (e) The ground truth.}
    \label{fig:ourtestsyn}
\end{figure}

\begin{table*}[ht]
  \centering
  \caption{PSNR/SSIM of our baseline deep models on different test sets. FOV, DM and ED denote the \textbf{FOV} distortion model, one-parameter \textbf{D}ivision \textbf{M}odel, and the \textbf{E}qui\textbf{D}istant distortion model, respectively.}
    \label{tab:baseline_psnr}
    
  \begin{threeparttable}
    \newcommand{\tabincell}[2]{\begin{tabular}{@{}#1@{}}#2\end{tabular}}
    \begin{tabular}{p{0.6cm}<{\centering}p{1.35cm}<{\centering}p{1.35cm}<{\centering}p{1.5cm}<{\centering}|p{1.35cm}<{\centering}p{1.35cm}<{\centering}p{1.5cm}<{\centering}|p{1.35cm}<{\centering}p{1.35cm}<{\centering}p{1.5cm}<{\centering}}
      \toprule
      {Dataset} & \multicolumn{3}{c}{\textbf{ADE20K}} & \multicolumn{3}{c}{\textbf{WireFrame}} & \multicolumn{3}{c}{\textbf{COCO}}\\
      \cmidrule{1-10}
      {} & FOV & DM & ED & FOV & DM & ED & FOV & DM & ED \\
      \cmidrule{1-10}
      FOV & \textbf{26.43/0.85}&16.65/0.45&18.84/0.54 & \textbf{26.45/0.86}&16.65/0.51&19.06/0.61 & \textbf{25.91/0.84}&16.09/0.43&18.51/0.53  \\
      DM & 21.03/0.63&24.76/0.81&\textbf{25.48/0.83}  & 21.02/0.68&25.10/0.83&\textbf{25.32/0.84} & 20.43/0.61&24.00/0.79&\textbf{24.83/0.81} \\
      ED & 18.83/0.56&23.37/0.75&\textbf{26.01/0.84}  & 19.02/0.63&23.77/0.79&\textbf{25.83/0.85} & 18.05/0.55&22.73/0.74&\textbf{25.45/0.83}  \\
      \bottomrule
    \end{tabular}
  \end{threeparttable}
\end{table*}

From Table \ref{tab:baseline_psnr}, we can find that the performance of each deep model is roughly the same across the three synthetic datasets that are based on the same distortion model, although it was only trained on the synthetic ADE20k dataset. For example, the DM ADE20k deep model achieves 24.76dB, 25.10dB, and 24.00dB on the DM ADE20k test dataset, DM WireFrame test dataset, and DM COCO test dataset respectively. The difference in the metrics among different source datasets is marginal. No matter if it is an image of indoor man-made furniture or a natural scene, our model could rectify the image since it had learned to know how the structural elements like lines should be corrected. These results imply that the key for image rectification is to find the distortion cues, e.g. lines, rather than the semantics of image contents. Since the performance of the deep models across the datasets is consistent, we did not train the deep models on the synthetic WireFrame and COCO datasets.

The other finding is that the ability to rectify various distorted images depends on the distortion model used in the network. For example, the FOV ADE20k model obtained 26.43 dB on the FOV ADE20k test set while the performance dropped significantly to 16.65 dB and 18.84 dB on the DM ADE20k and ED ADE20k test set, respectively. We can get the same observation from the subjective rectification results of the three deep models in Figure \ref{fig:ourtestsyn}. FOV ADE20k model obtained under-rectified results for some images synthesized based on the other two distortion models. By contrast, the DM ADE20k model could correct the distortions caused by the equidistant distortion model and FOV distortion model quite well. Indeed, the performance of the DM deep model was very stable among all the test sets. Generally, the DM deep model achieved the best average performance among all the models on all the test sets.

\begin{figure}[ht]
    \centering
    \includegraphics[width=1\linewidth]{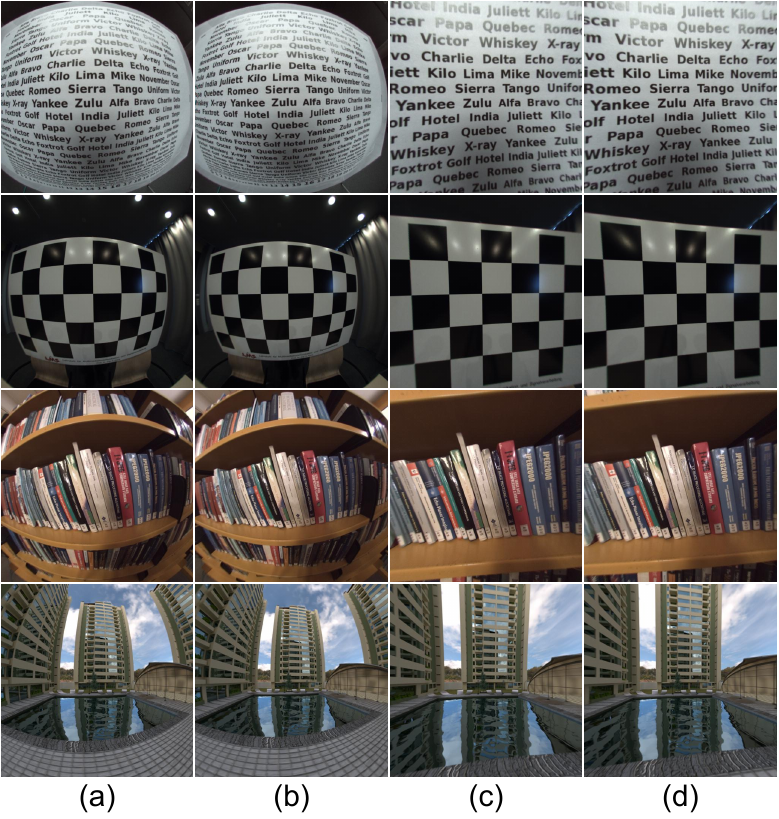}
    \caption{Results of our baseline models on real fisheye images \citep{eichenseer_DataSetProviding_2016}. (a) Input fisheye images. (b)-(d) Results of the FOV ADE20k model, the DM ADE20k model, and the ED ADE20k model. }
    \label{fig:ourtestreal}
\end{figure}

When tested on images from the real fisheye dataset \citep{eichenseer_DataSetProviding_2016}, the performance of the three deep models is also different from one to the other, as shown in Figure \ref{fig:ourtestreal}. We can easily find that both the DM deep model and ED deep model produced promising results, while the FOV deep model failed in most cases. The observation is the same as that in the synthesized dataset. The ability of one model to rectify the distortions generated by a model of its own family is called \textit{self-consistency} while the ability to rectify the distortions generated by other models outside of its family is called \textit{universality} \citep{tang_SelfconsistencyUniversalityCamera_2012}. From the results in Table~\ref{tab:baseline_psnr} and Figure~\ref{fig:ourtestsyn}, we can see that all three models are self-consistent, while the division model is more universal than the others. This conclusion is the same as that in \citep{tang_SelfconsistencyUniversalityCamera_2012}.

\subsection{Discussion} 
\begin{figure*}[ht]
    \centering
    \includegraphics[width=0.935\textwidth]{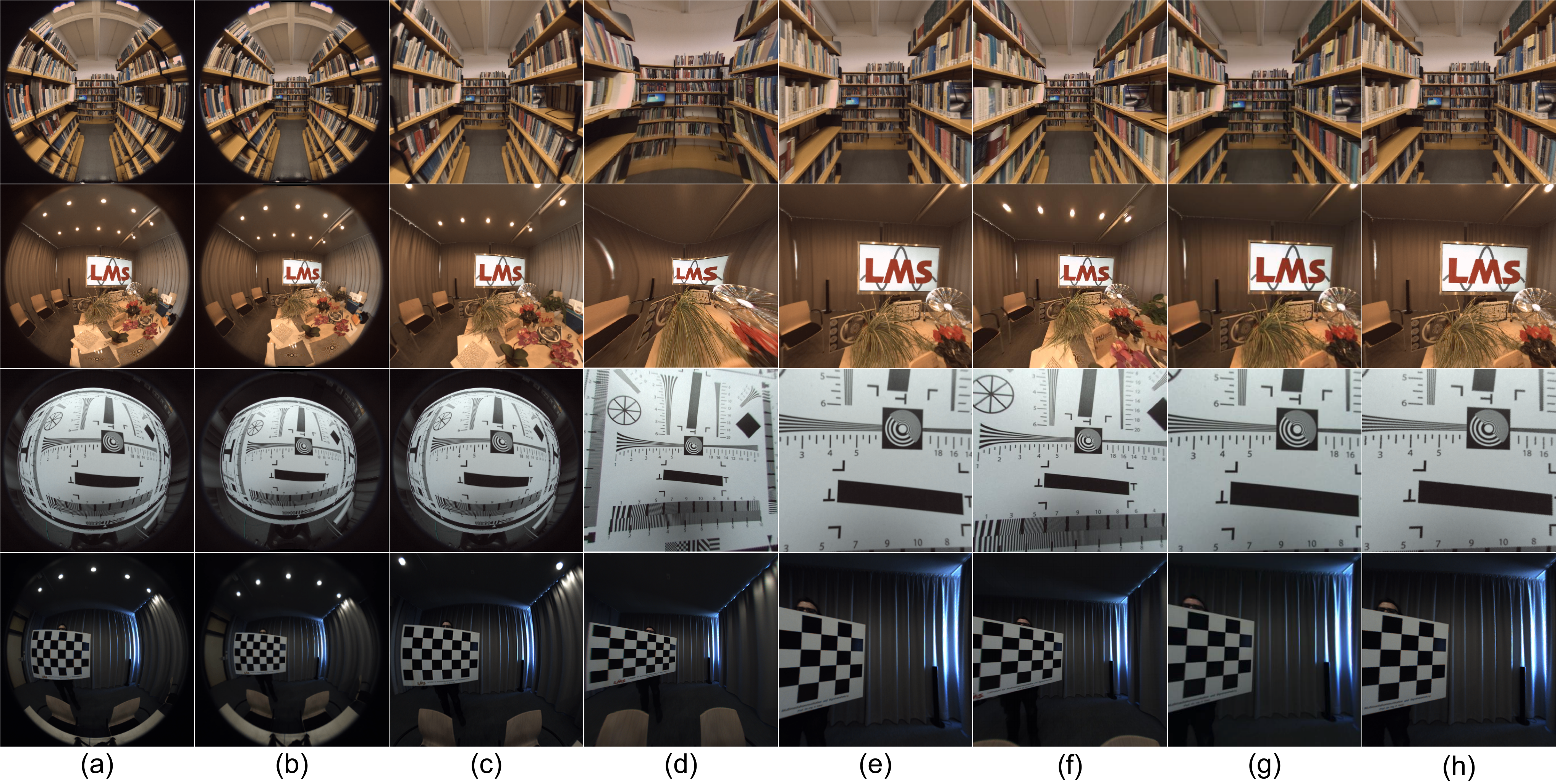}
    \caption{The results of SOTA methods and our baseline on images from the real fisheye dataset \citep{eichenseer_DataSetProviding_2016}. (a) The input fisheye images. (b) Results  of \cite{aleman-flores_AutomaticLensDistortion_2014}. (c) Results  of \cite{rong_RadialLensDistortion_2016}. (d) Results  of \cite{bogdan_DeepCalibDeepLearning_2018}. (e) Results  of \cite{xue_FisheyeDistortionRectification_2020}. (f) Results of Our baseline method using the DM ADE20k deep model. (g) The crop out results of (f). (h) The ground truth.}
    \label{fig:xuecomp}
\end{figure*}

In real-world applications, we want to find a universal distortion model \citep{tang_SelfconsistencyUniversalityCamera_2012} that can handle different types of distortions in real-world wide-angle images. However, from both the objective and subjective evaluation results of existing SOTA methods and our baseline models, we can see that it is difficult to obtain a universal model since each distortion model is based on specific assumptions and always adapted for a specific type of distortions \citep{liao_ModelFreeDistortionRectification_2020}. Owing to the powerful representation capability of deep neural networks, some methods try to incorporate multiple distortion models into one framework \citep{li_BlindGeometricDistortion_2019,liao_ModelFreeDistortionRectification_2020}, which makes it possible to provide a more general solution. Although multiple distortion models can improve the generalizability, they can not cover all the distortions in the real world. In Figure \ref{fig:xuecomp}, four SOTA methods and our baseline method were compared on images from the real fisheye dataset \citep{eichenseer_DataSetProviding_2016}. As can be seen, some of them failed on real fisheye images, e.g., \cite{rong_RadialLensDistortion_2016} obtained under-corrected results, while \cite{bogdan_DeepCalibDeepLearning_2018} obtained over-corrected results in some cases. Although \cite{xue_FisheyeDistortionRectification_2020} achieved promising results, it required rich lines to guide the rectification, which may not apply to images with fewer structures. Our baseline method obtained comparable results as \cite{xue_FisheyeDistortionRectification_2020}, but it was only trained on the synthetic dataset without the need for extra annotations. Besides, the methods in \cite{aleman-flores_AutomaticLensDistortion_2014} and \cite{bukhariAutomaticRadialDistortion2013} cannot handle these real fisheye neither.

\begin{table}[ht]
\scriptsize
    \centering
    \caption{Comparison of running time (seconds).}
    \label{tab:runtime}
    
  \newcommand{\tabincell}[2]{\begin{tabular}{@{}#1@{}}#2\end{tabular}}
    \begin{tabular}{ r l l}
    \toprule
    Methods & Platform & Time \\
    \midrule
    \cite{bukhariAutomaticRadialDistortion2013} & Intel i5-4200U CPU & 62.53  \\
    \cite{aleman-flores_AutomaticLensDistortion_2014} & Intel Xeon E5-1620 CPU & 2.13   \\
    \citep{zhang_LinebasedMultiLabelEnergy_2015} & Intel i5-4200U CPU & 80.07  \\
    \midrule
     \cite{rong_RadialLensDistortion_2016} & NVIDIA Tesla K80 GPU & 0.87  \\
    \cite{yin_FishEyeRecNetMulticontextCollaborative_2018} & NVIDIA Tesla K80 GPU & 1.31  \\
    \cite{liao_DRGANAutomaticRadial_2020}  & NVIDIA TITAN X GPU & 0.038  \\
    \midrule 
    Our baseline &  NVIDIA Tesla V100 GPU & \textbf{0.008}  \\
    \bottomrule
    \end{tabular}
\end{table}

Learning-based methods regress the parameters or estimate the warp field via a single forward-pass computation, no matter how complex the distortion model is or how many distortion models are involved. As for traditional geometry-based methods, we can also divide deep learning-based methods into one-stage methods and two-stage methods. If an independent post-processing step is needed to rectify the image using the estimated parameters or the warp field, the method is called a two-stage method, e.g. \citep{rong_RadialLensDistortion_2016}. If the rectification step is integrated into the deep network and the output is the corrected image, the method is called a one-stage method, e.g. \citep{yin_FishEyeRecNetMulticontextCollaborative_2018}. Generally, one-stage methods are faster than two-stage methods, but the performance still depends on the network capacity and the distortion model. In contrast to the learning-based methods, traditional methods often need to minimize a complex objective function iteratively, which is time-consuming and difficult to accelerate. Thereby, traditional methods are slower than learning-based methods in most cases and sometimes even 10-100 times slower. We collected the average running time of some traditional and learning-based methods from \citep{yin_FishEyeRecNetMulticontextCollaborative_2018} and \citep{liao_DRGANAutomaticRadial_2020} and summarized them in Table \ref{tab:runtime}. Although they were evaluated on different hardware, the results can still reveal the trend. From the table, we can see that all learning-based methods are faster than traditional methods, e.g., the one-stage learning-based method in \citep{liao_DRGANAutomaticRadial_2020} processed a $256 \times 256$ image in only 0.038 seconds. Our baseline method belongs to the one-stage method. We integrated the rectification layer in the deep model and generated the rectified image via a single forward-pass. We tested our method on the NVIDIA Tesla V100 GPU and it took 8 milliseconds to process a $257 \times 257$ image, i.e. 125 FPS, which is about $4\times$ faster than that of \citep{rong_RadialLensDistortion_2016}. It is noteworthy that although learning-based methods are always faster, a smaller image is usually used compared to that of the traditional methods, e.g. $256 \times 256$ in \citep{rong_RadialLensDistortion_2016, liao_DRGANAutomaticRadial_2020} and $257 \times 257$ in our baseline method. For model-based methods, the predicted parameters can be used to rectify high-resolution images directly with only more computations during warping. For the model-free methods, although the estimated warp field can be up-sampled to match the high resolution of the distorted image for warping, the details may be lost due to the up-sampling.

\section{Future Directions} \label{future}
Although existing methods have produced impressive results for certain types of distortions, there is currently no general solution for all distortion types. Furthermore, as the number of regulation terms in the objective function increases, the computational complexity and optimization stability become intractable, making it difficult to deal with various types of distortions. Due to the strong representation capacity of deep neural networks, deep learning-based methods have become popular and are delivering promising results. Nevertheless, more effort is needed to improve overall performance. We discuss several promising future research directions below.

\textbf{Distortion Model-independent Rectification.} In both traditional geometry-based methods and deep learning-based methods, specific distortion models are used to model the distortion explicitly or implicitly. However, these can only represent certain distortion types, thereby limiting the application of these rectification methods. Although efforts have been made to utilize several models at the same time \citep{li_BlindGeometricDistortion_2019,liao_ModelFreeDistortionRectification_2020}, the included distortion types are still too limited to account for all the distortion types found in real-world wide-angle images. The distortion model can be seen as a bridge between the distorted image domain and the normal image domain, through which constraints or supervision can be constructed, e.g., straight lines in normal images become circular arcs in distorted images under the one-parameter division model \citep{brauer-burchardt_NewAlgorithmCorrect_2001}. If we have prior knowledge about what the objects should look like in the scene (e.g., a wall being vertical and the projection of a ball being a circle), new losses based on it can be used to guide the rectification or to supervise training, negating the need for a distortion model.

\textbf{Unpaired Training Data.} Existing deep learning-based methods train the network using distorted and undistorted image pairs, which are hard to collect from the real world. Alternatively, they can be synthesized based on some specific and limited distortion models. Most of these methods define the rectification problem as a regression of the distortion parameters or an estimate of the warp field derived from the distortion model. If the distorted and undistorted images are regarded as samples from two different domains, image rectification can be formulated as an unsupervised or self-supervised image-to-image translation problem, in which paired training data may not be necessary \citep{zhu_UnpairedImagetoImageTranslation_2017, chao_SelfSupervisedDeepLearning_2020, fan2020sir}. In this case, different consistency constraints could be explored, e.g., cycle consistency and geometric constraints of structural elements. Compared to the image style or texture transfer tasks \citep{gatys_ImageStyleTransfer_2016,isola_ImagetoImageTranslationConditional_2017}, image rectification is restricted by the geometric consistency of the image contents.

\textbf{Perceptual Quality Assessment.} Not all lines and shapes can be simultaneously preserved for wide-angle image rectification. There needs to be a trade-off between different distortion terms to find a feasible solution that favors specific aspects, which is subjective in nature. People may give a significantly different quality assessment for the same image conditioned on their perceptual preferences. Distinct, subjective metrics have been used to measure image quality in various tasks. Therefore, one can use the perceptual image quality assessment metric to guide rectification, such that even if the rectified image is not the same as the ground truth undistorted image, it has a better perceptual quality. Moreover, the attention mechanism can play an important role in this kind of subjective evaluation metric, which is also worthy of further study.

\textbf{High-resolution Image Rectification.}  Almost all existing deep learning-based methods are trained on low-resolution images, i.e., typically smaller than $350 \times 350$. Since high-resolution images have now become very common as camera sensors have improved, high-resolution image rectification is important in practice. However, it faces two major obstacles: the computational cost and the recovery of details in regions far from the distortion center. While the former can be addressed by designing lightweight neural networks and leveraging modern GPUs for acceleration, the latter is an inherently challenging problem due to the inhomogeneous resolution of distorted images. Borrowing ideas from the areas of image super-resolution \citep{ledig_PhotoRealisticSingleImage_2017, lim_EnhancedDeepResidual_2017, wang_DeepLearningImage_2020} and inpainting \citep{bertalmio_ImageInpainting_2000, bertalmio_SimultaneousStructureTexture_2003, yu_GenerativeImageInpainting_2018, elharrouss_ImageInpaintingReview_2020} may be helpful to address this issue. 

\textbf{Loss Functions.} In existing methods, a typical loss is calculated as the difference between the original image and the predicted image, i.e., L1 or L2 loss, which is the key component of the total objective function. In unsupervised or self-supervised training methods, where the models are trained with unpaired training images, new losses should be carefully designed to preserve the structural elements and salient contents of images. Perceptual losses are also worth exploring to guide the rectification model to generate a visually pleasing result.

\textbf{Benchmark Datasets.}  Almost all deep learning-based methods use their own synthetic training and test sets, which are synthesized based on different distortion models with different parameters. Since rectification model performance depends on the training data, it is hard to disentangle each method’s performance from the specific synthetic dataset used. Therefore, it is crucial to establish a benchmark dataset containing both real-world and synthetic images with various types of distortions as well as annotations to evaluate and compare different methods using the same protocol.

\section{Conclusion} \label{conclusion}
In this paper, we present a comprehensive survey of progress in the area of wide-angle image rectification. Some typical camera models and distortion models playing a fundamental role in image rectification are described and discussed. We empirically find that the division model has the best universality. Models trained on synthetic data have the best generalizability to both synthetic images with other types of distortions and real-world fisheye images. Moreover, we comprehensively review progress in two main types of image rectification methods, i.e., traditional geometry-based methods and deep learning-based methods. Specifically, we discuss their relationships, differences, strengths, and limitations. We also evaluate the performance of state-of-the-art methods on public synthetic and real-world datasets. Generally, deep learning-based methods are promising approaches that merit further study, achieving good performance and running faster than traditional geometry-based methods. We also devise a new baseline model that has comparable performance with SOTA methods. Some ongoing challenges and potential research directions in this area are also summarized. We hope that this survey benefits future research on this topic.

\bibliographystyle{apalike}
\bibliography{image_rectification_revision.bib}

\end{document}